\documentclass{article}

\usepackage{microtype}
\usepackage{graphicx}
\usepackage{multirow}
\usepackage{wrapfig}
\usepackage{subfig}
\usepackage{booktabs} % for professional tables
\usepackage[hyphens]{url}
\usepackage{caption}
\usepackage{enumitem}

\usepackage{hyperref}
\usepackage{wrapfig}

\usepackage[accepted]{icml2024/icml2024}

\usepackage{amsmath}
\usepackage{amssymb}
\usepackage{mathtools}
\usepackage{amsthm}
\mathtoolsset{showonlyrefs} 

\usepackage[capitalize,noabbrev]{cleveref}
	
\theoremstyle{plain}
\newtheorem{theorem}{Theorem}[section]

\newtheorem{corollary}[theorem]{Corollary}
\theoremstyle{definition}
\newtheorem{definition}[theorem]{Definition}
\newtheorem{assumption}[theorem]{Assumption}
\theoremstyle{remark}

\DeclareMathOperator*{\argmax}{arg\,max}
\DeclareMathOperator*{\argmin}{arg\,min}
\DeclarePairedDelimiterX{\infdivx}[2]{(}{)}{%
	#1\;\delimsize\|\;#2%
}
\newcommand{\kl}{\textrm{KL}\infdivx}

\usepackage{comment}

\graphicspath{{Figures/}}
\icmltitlerunning{Bayesian Exploration Networks}

\begin{document}
\twocolumn[
\icmltitle{Bayesian Exploration Networks}



\icmlsetsymbol{equal}{*}

\begin{icmlauthorlist}
	\icmlauthor{Mattie Fellows}{equal,oxeng}
	\icmlauthor{Brandon Kaplowitz}{equal,nyu}
	\icmlauthor{Christian Schroeder de Witt}{oxeng}
	\icmlauthor{Shimon Whiteson}{oxcomp}
\end{icmlauthorlist}
\icmlaffiliation{oxeng}{Department of Engineering Science, University of Oxford, Oxford, United Kingdom}
\icmlaffiliation{nyu}{Department of Economics, New York University, New York, United States of America}
\icmlaffiliation{oxcomp}{Department of Computer Science, University of Oxford, Oxford, United Kingdom}

\icmlcorrespondingauthor{Mattie Fellows}{matthew.fellows@eng.ox.ac.uk}

\icmlkeywords{Machine Learning, ICML}

\vskip 0.3in
]



\printAffiliationsAndNotice{\icmlEqualContribution} % otherwise use the standard text.

\begin{abstract}
	Bayesian reinforcement learning (RL) offers a principled and elegant approach for sequential decision making under uncertainty. Most notably, Bayesian agents do not face an exploration/exploitation dilemma, a major pathology of frequentist methods. However theoretical understanding of model-free approaches is lacking. In this paper, we introduce a novel Bayesian model-free formulation and the first analysis showing that model-free approaches can yield Bayes-optimal policies. We show all existing model-free approaches make approximations that yield policies that can be arbitrarily Bayes-suboptimal. As a first step towards model-free Bayes optimality, we introduce the Bayesian exploration network   (\textsc{ben}) which uses normalising flows to model both the aleatoric uncertainty (via density estimation) and epistemic uncertainty (via variational inference) in the Bellman operator. In the limit of complete optimisation, \textsc{ben} learns true Bayes-optimal policies, but like in variational expectation-maximisation, partial optimisation renders our approach tractable. Empirical results demonstrate that \textsc{ben} can learn true Bayes-optimal policies in tasks where existing model-free approaches fail.
\end{abstract}
\vspace{-0.5cm}\section{Introduction}
\label{sec:introduction}
In reinforcement learning (RL), an agent is tasked with learning an optimal policy that maximises expected return in a Markov decision process (MDP). In most cases, the agent is in a learning setting and does not know the underlying MDP a priori: typically, the reward and transition distributions are unknown. A Bayesian approach to reinforcement learning characterises the uncertainty in unknown governing variables in the MDP by inferring a posterior over their values conditioned on observed histories of interactions. Using the posterior it is possible to marginalise across unknown variables and derive a belief transition distribution that characterises how the uncertainty will evolve over all future timesteps. The resulting Bayesian RL (BRL) objective transforms a learning problem into a planning problem with a well defined set of optimal policies, known as Bayes-optimal policies, which are a gold standard for exploration \citep{Martin67,Duff02}. \iffalse From this perspective, the exploration/exploitation dilemma is a major pathology of frequentist RL due to the violation of the \emph{conditionality principle}: when in a learning problem, frequentist methods can condition on information that the agent does not have access to, namely the unknown transition and reward distributions. Frequentist RL must close this gap with exploration heuristics as there is no formal method to tackling this dilemma. By contrast, Bayes-optimal policies solve the exploration/exploitation dilemma by exploring to reduce epistemic uncertainty in the MDP, but only insofar as doing so increases expected returns as the belief evolves across timesteps.\fi  
Moreover, any non-Bayesian policy is suboptimal in terms of optimising the expected returns according to the belief induced by the prior and model of the state and reward transition distributions. 

Despite the formal theoretical benefits, learning Bayes-optimal policies remains a significant challenge due to several sources of intractability. Firstly, model-based approaches must maintain a posterior over a model of the state transition dynamics, which is notoriously computationally complex for even low dimensional state spaces \citep{Wasserman06}. Secondly, even if it is tractable to calculate and maintain the posterior, the marginalisation needed to find the Bayesian transition and reward distributions requires high dimensional integrals. Finally, given the Bayesian distributions, a planning problem must then be solved in belief space for every history-augmented state to obtain the Bayes-optimal policy.

Alternatively, model-free approaches characterise uncertainty in a Bellman operator. This avoids the issues of modelling uncertainty in high dimensional transition distributions, as Bellman operators require the specification of a one-dimensional conditional distribution. Whilst model-free approaches to BRL exist, little is known about their theoretical properties. Our main contribution is to provide a theoretical analysis to answer three core questions: I) Can model-free approaches learn Bayes-optimal policies? II) What are the relative benefits of model-free approaches? and III) Are existing model-free approaches Bayes-optimal? We answer these questions by introducing a novel model-free formulation that is provably Bayes-optimal whilst still characterising uncertainty in a low-dimensional distribution over Bellman operators. Moreover, we prove that all existing methods inadvertently solve an approximation to the true Bayesian objective that prevents them from learning a true Bayes-optimal policy. 

 Motivated by our analysis, we introduce a Bayesian exploration network (\textsc{ben}) for model-free BRL. \textsc{ben} reduces the dimensionality of inputs to a one-dimensional variable using a $Q$-function approximator. The output is then passed through a Bayesian network. Like in an actor-critic approach, \textsc{ben} can be trained using partial stochastic gradient descent (SGD) methods at each timestep, bypassing computational complexity issues associated with finding a Bayes-optimal policy. This comes at the expense of learning an approximately Bayes-optimal policy but one that converges to the true Bayes-optimal policy in the limit of complete optimisation. To verify our theoretical claims, we evaluate \textsc{ben} in a search and rescue environment, which is a novel higher dimensional variant of the tiger problem \citep{Kaelbling98}. We show \textsc{ben} solves the task while oracles of existing state-of-the-art model-free BRL approaches based on BootDQN+Prior \citep{Osband18} and Bayesian Bellman Actor Critic \citep{Fellows21} fail due to their inability to learn Bayes-optimal policies. Moreover, our results show that, whilst in the limit of complete optimisation \textsc{ben} recovers true Bayes-optimal policies, complete optimisation is not necessary as \textsc{ben} behaves near Bayes-optimally after taking only a few optimisation steps on our objective for every observation. 

 \vspace{-0.2cm}
 \section{Preliminaries}
  \vspace{-0.1cm}
\subsection{Contextual RL}\label{sec:contextual_rl}  \vspace{-0.1cm}
 We define a space of infinite-horizon, discounted contextual Markov decision processes (CMDPs) \citep{Hallak2015} by introducing a context variable $\phi\in\Phi\subseteq \mathbb{R}^d$:  $\mathcal{M}(\phi)\coloneqq \langle \mathcal{S},\mathcal{A}, P_0, P_S( s,a,\phi), P_R( s,a,\phi), \gamma \rangle$ where each $\phi$ indexes a specific MDP by parametrising a transition distribution $P_\mathcal{S}( s,a,\phi):\mathcal{S}\times\mathcal{A}\times\Phi\rightarrow \mathcal{P}(\mathcal{S})$ and reward distribution $P_R( s,a,\phi):\mathcal{S}\times\mathcal{A}\times\Phi\rightarrow \mathcal{P}(\mathbb{R})$. We denote the corresponding joint conditional state-reward transition distribution as $P_{R,S}(s,a,\phi)$. We assume that the agent has complete knowledge of the set of states $\mathcal{S}\subseteq \mathbb{R}^n$, set of actions $\mathcal{A}$, initial state distribution $P_0\in \mathcal{P}(\mathcal{S})$ and discount factor $\gamma$. An agent follows a policy $\pi:\mathcal{S}\times \Phi \rightarrow \mathcal{P}(\mathcal{A})$, taking actions $a_t\sim \pi(s_t,\phi)$. We denote the set of all context-conditioned policies as $\Pi_\Phi\coloneqq \{\pi: \mathcal{S}\times\Phi\rightarrow\mathcal{P}(\mathcal{A}) \}$. 
The agent is assigned an initial state $s_0\sim P_0$. As the agent interacts with the environment, it observes a history of data  $h_t\coloneqq \{s_0,a_0,r_0, s_1,a_1,r_1,\ldots a_{t-1},r_{t-1},s_t\}\in\mathcal{H}_t$ where $\mathcal{H}_t$ is the corresponding state-action-reward product space. We denote the context-conditioned distribution over history $h_t$ as: $P^\pi_{t}( \phi)$ with density $p^\pi_{t}(h_t\vert \phi)=p_0(s_0)\prod_{i=0}^t \pi(a_i\vert s_i,\phi)p(r_i,s_{i+1}\vert s_i,a_i,\phi)$.

In the infinite-horizon, discounted setting, the goal of an agent in MDP $\mathcal{M}(\phi)$ is to find a policy that optimises the objective: $J^\pi(\phi)=\mathbb{E}_{\tau_\infty \sim P^\pi_\infty( \phi)}\left[\sum_{t=0}^\infty \gamma^t r_t\right]$. We denote an optimal policy as $\pi^\star(\cdot,\phi)\in\Pi^\star_\Phi(\phi)\coloneqq \argmax_{\pi\in \Pi_\Phi} J^\pi(\phi)$,
where $\Pi^\star_\Phi(\phi)$ is the set of all optimal MDP-conditioned policies that are optimal for $\mathcal{M}(\phi)$. For an optimal policy $\pi^\star$, the optimal quality function ($Q$-function) $ Q^\star:\mathcal{S}\times \mathcal{A}\times \Phi \rightarrow \mathbb{R}$ satisfies the optimal Bellman equation: $	\mathcal{B}^\star \left[Q^\star\right](s_t,a_t,\phi)=Q^\star(s_t,a_t,\phi)$ where  
\begin{align}
	&\mathcal{B}^\star \left[Q^\star\right](s_t,a_t,\phi)\\
	&\coloneqq \mathbb{E}_{r_t,s_{t+1}\sim P_{R,S}( s_t,a_t,\phi) }[r_t+  \max_{a'\in\mathcal{A}}Q^\star (s_{t+1},a',\phi)],
\end{align}
is the optimal Bellman operator.

If the agent has access to the true MDP $\mathcal{M}(\phi^\star)$, computational complexity issues aside, an optimal policy can be obtained by solving a \emph{planning} problem \iffalse either by optimising the RL objective $J^\pi(\phi^\star)$ directly for $\pi$ or by solving an optimal Bellman equation and taking the action $a_t\in\argmax_{a'\in\mathcal{A}}Q^\star(s_t,a',\phi^\star)$\fi. In the more realistic setting, the agent does not have access to the MDP's transition dynamics and/or reward function. The agent must balance learning these variables through exploration of the MDP at the cost of behaving suboptimally with solving the underlying planning problem by exploiting the information it has observed. This setting is known as a \emph{learning} problem and solving the exploration/exploitation dilemma remains a major challenge for any agent learning to behave optimally. 
		\vspace{-0.5cm}
\subsection{Bayesian RL}\label{sec:Bayesian_rl}		\vspace{-0.1cm}
A Bayesian epistemology characterises the agent's uncertainty in the MDP through distributions over $\Phi$. We start by defining the prior distribution $P_\Phi$ which represents the a priori belief in the true value $\phi^\star$ before the agent has observed any transitions. Priors are a powerful aspect of BRL, allowing practitioners to provide the agent with any information about the MDP and transfer knowledge between agents and domains. In the tabula rasa setting, priors can be uninformative, can be used to encode optimism or pessimism in unknown states; or can be a minimax prior representing the worst possible prior distribution over MDPs an agent could face \citep{Buening23}. Given a history $h_t$, we aim to reason over future trajectories; thus, Bayesian agents follow policies that condition on histories rather than single states. We denote the space of all histories $\mathcal{H}\coloneqq \{\mathcal{H}_t\vert t\ge 0\}$ and the set of all history-conditioned policies as $\Pi_\mathcal{H}\coloneqq \{\pi: \mathcal{H}\rightarrow\mathcal{P}(\mathcal{A}) \}$. A Bayesian agent characterises the uncertainty in the MDP by inferring the posterior $P_\Phi(h_t)$ for each $t\ge0$.

The prior is a special case of the posterior with $h_t=\varnothing$. The posterior $P_\Phi(h_t)$ represents the agent's beliefs in the MDP and can be used to \emph{marginalise} across all CMDPs according to the agent's uncertainty. This yields the Bayesian state-reward transition distribution: $P_{R,S}(h_t,a_t)\coloneqq\mathbb{E}_{\phi\sim P_\Phi( h _t)} \left[P_{R,S}(s_{t},a_{t},\phi)\right] $. Given this distribution, we can reason over counterfactual future trajectories using the prior predictive distribution over trajectories $P_{t}^\pi$ with density: $p_{t}^\pi(h_t)= p_0(s_0) \prod_{i=0}^{t}\pi(a_i\vert h_i)p(r_i,s_{i+1}\vert h_i,a_i)$. Using the predictive distribution, we define the BRL objective as $J^\pi_\textrm{Bayes}\coloneqq \mathbb{E}_{h_\infty \sim P_{\infty}^\pi}\left[ \sum_{i=0}^\infty \gamma^{i} r_{i} \right]$. A corresponding optimal policy is known as a Bayes-optimal policy, which we denote as $\pi^\star_{\textrm{Bayes}}(\cdot)\in \Pi^\star_{\textrm{Bayes}}\coloneqq \argmax_{\pi\in\Pi_\mathcal{H}} J^\pi_\textrm{Bayes}$.

 Unlike in frequentist RL, Bayesian variables depend on histories obtained through posterior marginalisation; hence the posterior is often known as the \emph{belief state}, which augments each ground state $s_t$ like in a partially observable MDP (POMDP).  Analogously to the state-transition distribution in frequentist RL, we can define a belief transition distribution $P_\mathcal{H}(h_t,a_t)$ using the Bayesian reward and transition distributions, which has the density:  $p_\mathcal{H}(h_{t+1}\vert h_t,a_t)= p(s_{t+1},r_t\vert h_t,a_t) \underbrace{p( h_t,a_t\vert  h_t,a_t)}_{=1}
 	=p(s_{t+1},r_t\vert h_t,a_t)$.

 Using the belief transition, we define the \emph{Bayes-adaptive MDP} (BAMDP) \citep{Duff02}:
 \begin{align}
 \mathcal{M}_\textrm{BAMDP}\coloneqq \langle\mathcal{H}, \mathcal{A}, P_0, P_\mathcal{H}(h,a), \gamma \rangle,
 \end{align}
 which can be solved using planning methods to obtain a Bayes-optimal policy \citep{Martin67}.

A Bayes-optimal policy naturally balances exploration with exploitation: after every timestep, the agent's uncertainty is characterised via the posterior conditioned on the history $h_{t}$, which includes all future trajectories to marginalise over. The BRL objective, therefore, accounts for how the posterior evolves after each transition, and hence any Bayes-optimal policy $\pi^\star_\textrm{Bayes}$ is optimal not only according to the epistemic uncertainty at a single timestep but also to the epistemic uncertainty at every future timestep, decaying according to the discount factor. 

Unlike in frequentist RL, if the agent is in a learning problem, finding a Bayes-optimal policy is always possible given sufficient computational resources. This is because any uncertainty in the MDP is marginalised over according to the belief characterised by the posterior. BRL thus does not suffer from the exploration/exploitation dilemma as actions are sampled from optimal policies that only condition on historical observations $h_t$, rather than the unknown MDP $\phi^\star$. More formally, this is a direct consequence of the \emph{conditionality principle}, which all Bayesian methods adhere to, meaning that Bayesian decisions never condition on data that the agent has not observed. From this perspective, the exploration/exploitation dilemma is a pathology that arises because frequentist approaches violate the conditionality principle. 

For a Bayes-optimal policy $\pi^\star$, we define the optimal Bayesian $Q$-function as $Q^\star(h_t,a_t)\coloneqq Q^{\pi^\star_\textrm{Bayes}}(h_t,a_t)$, which satisfies the optimal Bayesian Bellman equation $Q^\star(h_t,a_t)=\mathcal{B}^\star[Q^\star](h_t,a_t)$ where: 
\begin{align}
	&\mathcal{B}^\star[Q^\star](h_t,a_t)\\
	&\coloneqq\mathbb{E}_{h_{t+1}\sim P_\mathcal{H}(h_t,a_t)}\left[r_t+\gamma \max_{a'}Q^\star(h_{t+1},a')\right],
\end{align}
is the optimal Bayesian Bellman operator. It is possible to construct a Bayes-optimal policy by choosing the action that maximises the optimal Bayesian $Q$-function $a_t\in\argmax_{a'} Q^\star(h_t,a')$; hence learning $Q^\star(h_t,\cdot)$ is sufficient for solving the BAMDP. We take this value-based approach in this paper.

\vspace{-0.2cm}

\subsection{Related Work}\vspace{-0.1cm}
\label{sec:related}
\textsc{ben} is the first model-free approach to BRL that can learn Bayes-optimal policies. To relate \textsc{ben} to other approaches, we clarify the distinction between model-free and model-based BRL: 

\begin{definition}
	Model-based approaches define a prior $P_\Phi$ over and a model of the MDP's state and reward transition distributions: $P_S(s,a,\phi)$ and $P_R(s,a,\phi)$. Model-free approaches define a prior $P_\Phi$ over and a model of the MDP's Bellman operators: $P_B(\cdot,\phi)$.\vspace{-0.1cm}
\end{definition} 
This definition mirrors classical interpretations of model-based and model-free RL, which categorises algorithms according to whether a model of transition dynamics is learnt or the $Q$-function is estimated directly \citep{Sutton18}. We formulate a novel Bayesian model-free approach in \cref{sec:model-free-formulation}  before proving in \cref{proof:model-free_model-based} that whichever approach is taken, a Bayes-optimal policy may still be learnt. 

We provide a review of several model-based approaches and their approximations in  \cref{app:model-based_approaches}, focusing instead on model-free approaches here. The majority of existing model-free approaches purportedly infer a posterior over $Q$-functions $P_Q^\pi(h_t)$ given a history of samples $h_t$, thus requiring a model of the aleatoric uncertainty in $Q$-function samples $q\sim P_Q^\pi( s,a, \phi)$. $P_Q^\pi( s,a, \phi):\mathcal{S}\times\mathcal{A}\times \Phi\rightarrow \mathcal{P}(\mathbb{R})$ is typically a parametric Gaussian, which is a conditional distribution over a one-dimensional space, allowing for standard techniques from Bayesian regression to be applied. As inferring a posterior over $Q$-functions requires samples from complete returns, some degree of bootstrapping using function approximation is required for algorithms to be practical \citep{Kuss03,Engel05,Gal16a,Osband18,Fortunato18,Lipton18,Osband19,Touati19}. By introducing bootstrapping, model-free approaches actually infer a posterior over \emph{Bellman operators}, which concentrates on the true Bellman operator with increasing samples under appropriate regularity assumptions \citep{Fellows21}. Instead of attempting to solve the BAMDP exactly, existing model-free approaches employ posterior sampling where a single MDP is drawn from the posterior at the start of each episode \citep{Thompson33,Strens00,Osband13}, or optimism in the face of uncertainty (OFU) \citep{Lai85,Kearns02} where exploration is increased or decreased by a heuristic to reflect the uncertainty characterised by the posterior variance \citep{Chi18,ciosek2019better,Luis23}.  Unfortunately, both
posterior sampling and  OFU exploration can be highly inefficient and far from Bayes-optimal \citep{Zintgraf19,Buening23}. Exploration strategies aside, a deeper issue with existing model-free Bayesian approaches is that an optimal policy under their formulations is not Bayes-optimal, but instead solves either a \emph{myopic} or \emph{QBRL} approximation to the BRL objective. We explore these approximations theoretically in \cref{sec:qbrl,sec:myopic} respectively. 
\vspace{-0.3cm}
\section{Analysis of Model-Free BRL}\vspace{-0.1cm}
\emph{Detailed proofs for theorems are provided in \cref{app:proofs}.}
We now provide an analysis of model-free BRL under function approximation to answer the following questions: 	\vspace{-0.3cm}
\begin{enumerate}[label=\Roman*)]
	\item Can model-free BRL approaches yield Bayes-optimal solutions and is there an analytic relationship between model-based and model-free approaches?
	\item What are the benefits of model-free approaches over model-based approaches?
	\item Are existing model-free BRL approaches Bayes-optimal and, if not, what solutions do these approaches learn instead? 
\end{enumerate}\vspace{-0.3cm}
 We introduce a parametric function approximator $Q_\omega: \mathcal{H}\times \mathcal{A}\rightarrow \mathbb{R}$ parametrised by $\omega\in\Omega$ to approximate the optimal Bayesian $Q$-function. The function approximator satisfies the following regularity assumption:
\begin{assumption}\label{ass:measurable} Assume that $Q_{\omega}(h_t,a_t)$ is a Lebesgue measurable mapping $Q_{\omega}( h_t\setminus s_t, a_t,\cdot):\mathcal{S}\rightarrow \mathbb{R}$ for all $\omega\in\Omega, h_t\setminus s_t\in\mathcal{H}$ and $a_t\in\mathcal{A}$ and the set $\argmax_{a'} Q_{\omega}( h_t, a')$ always exists.
\end{assumption}\vspace{-0.2cm}
\cref{ass:measurable} should automatically be satisfied for most function approximators of interest in RL (for example bounded Lipschitz functions defined on closed sets). 
\vspace{-0.2cm}
\subsection{Model-Free BRL Formulation}\vspace{-0.1cm}
\label{sec:model-free-formulation}
In model-free BRL, our goal is to characterise uncertainty in the optimal Bayesian Bellman operator instead of the reward-state transition distribution. Given samples from the true reward-state distribution $r_t,s_{t+1} \sim P^\star_{R,S}(s_t,a_t)$ we use  \emph{bootstrapping} to estimate the optimal Bayesian Bellman operator
\begin{align}
	b_t= \beta_\omega(h_{t+1})\coloneqq r_t +\gamma\max_{a'} Q_\omega(h_{t+1},a').\vspace{-0.5cm}
\end{align}
We refer to $\beta_\omega(h_{t+1})$ as the bootstrap function. Similarly to \citet{Fellows21}, we interpret bootstrapping as making a change of variables under the mapping $\beta_\omega(\cdot,h_t,a_t):\mathbb{R}\times \mathcal{S}\rightarrow \mathbb{R}$. The bootstrapped samples $b_t$ have distribution $P_B^\star(h_t,a_t;\omega)$ which is the \emph{pushforward} distribution over next period's possible updated Q-values satisfying: $	\mathbb{E}_{b_t\sim P_B^\star(h_t,a_t;\omega) }\left[ f(b_t) \right]=\mathbb{E}_{r_t,s_{t+1}\sim P_{R,S}^\star(s_t,a_t) }\left[ f\left(r_t+\gamma\max_{a'} Q_\omega(h_{t+1},a')
\right)\right]$ for any measurable function $f:\mathbb{R}\rightarrow \mathbb{R}$. We refer to $P_B^\star(h_t,a_t;\omega)$ as the Bellman distribution.

When predicting $b_t$ given an observation $h_t,a$, there are two sources of uncertainty to take into account: firstly, even if $P_B^\star(h_t,a_t;\omega)$ is known, there is natural stochasticity due to the environment's reward-state transition dynamics that prevents $b_t$ being determined. This type of uncertainty is known as \emph{aleatoric uncertainty}. Secondly, in a learning problem, the Bellman distribution $P_B^\star(h_t,a_t;\omega)$ cannot be determined a priori and must be inferred from observations of $b_t$. This type of uncertainty is known as \emph{epistemic uncertainty}. Unlike aleatoric uncertainty, epistemic uncertainty can always be reduced with more data as the agent explores.

 Our first step is to introduce a model of the process $b_t\sim P_B^\star(h_t,a_t;\omega)$, which characterises the aleatoric uncertainty in the optimal Bellman operator. In this paper, we choose a parametric model $P_B(h_t,a_t,\phi;\omega)$ with density $p(b_t\vert h_t,a_t,\phi;\omega)$ parametrised by $\phi\in\Phi$; however a nonparametric model may also be specified. We specify a prior $P_\Phi$ over the parameter space which represents the agent's initial belief over the space of models. The space of models $P_B(h_t,a_t,\phi;\omega)$ can be interpreted as a hypothesis space over the true Bellman distribution $P_B^\star(h_t,a_t;\omega)$, with each hypothesis indexed by a parameter $\phi\in\Phi$.

 Let $\mathcal{D}_\omega(h_t)\coloneqq \{(b_i,h_i,a_i)\}_{i=0}^{t-1}$ denote the dataset of bootstrapped samples. Once the agent has observed $\mathcal{D}_\omega(h_t)$, it updates its belief in $\phi$ by inferring a posterior $P_\Phi(\mathcal{D}_\omega(h_t))$, whose density can be derived using Bayes' rule: 
 \begin{align}
 p(\phi\vert \mathcal{D}_\omega(h_t))=\frac{\prod_{i=0}^{t-1} p(b_i\vert h_i,a_i,\phi;\omega)p(\phi)}{\mathbb{E}_{\phi\sim P_\Phi}\left[\prod_{i=0}^{t-1} p(b_i\vert h_i,a_i,\phi;\omega)\right]}. 
 \end{align}
The posterior $P_\Phi(\mathcal{D}_\omega(h_t))$ characterises the epistemic uncertainty over the hypothesis space, which we use to obtain the predictive optimal Bellman distribution: $P_B(h_t,a_t;\omega)= \mathbb{E}_{\phi\sim P_\Phi ( \mathcal{D}_\omega(h_t))}\left[P_B(h_t,a_t,\phi;\omega)\right].$
 
  The predictive optimal Bellman distribution is analogous to the predictive reward-state transition distribution introduced in \cref{sec:Bayesian_rl} as it is derived by marginalising across the hypothesis space according to the epistemic uncertainty in each model under the posterior. Taking expectations over the variable $b_t$ using $	P_B(h_t,a_t;\omega)$, we derive the predictive optimal Bellman operator: $B^+[Q_\omega](h_t,a_t )\coloneqq\mathbb{E}_{b_t\sim P_B(h_t,a_t;\omega)}\left[b_t\right]$,
which integrates both the aleatoric and epistemic uncertainty in $b_t$ to make a Bayesian prediction of the optimal Bellman operator at each timestep $t$. Intuitively, we expect $B^+[Q_\omega](h_t,a_t )$ to play the same role as the optimal Bayesian Bellman operator introduced in \cref{sec:Bayesian_rl}, which we now formally confirm:
\begin{theorem} \label{proof:model-free_model-based} 
	Let \cref{ass:measurable} hold, then $B^+[Q_\omega](h_t,a_t )=B^\star[Q_\omega](h_t,a_t )$.
\end{theorem}  
\vspace{-0.1cm}
 In answer to Question I, \cref{proof:model-free_model-based} proves that model-free approaches can be Bayes-optimal: if $Q_{\omega^\star}$ satisfies $Q_{\omega^\star}(\cdot)=B^+[Q_{\omega^\star}](\cdot)$, it also satisfies an optimal Bayesian Bellman equation: $Q_{\omega^\star}(\cdot)=\mathcal{B}^\star[Q_{\omega^\star}](\cdot)$ hence any agent taking action $a_t \in \argmax_{a'} Q_{\omega^\star}(h_t,a')$ is thus Bayes-optimal with respect to the prior $P_\Phi$ and likelihood defined by the model $P_B(h_t,a_t,\phi;\omega^\star)$.  \cref{proof:model-free_model-based} is a consequence of the \emph{sufficiency principle}: this result confirms that it does not matter whether we characterise uncertainty in the reward-state transition distributions or the optimal Bellman operator, a Bayes-optimal policy may still be learned. 
 \vspace{-0.2cm}
 \subsection{Aleatoric Uncertainty Matters}
 \label{sec:aleatoric}
\begin{figure}\vspace{-0.1cm}
	\centering
	\includegraphics[width=0.4\textwidth]{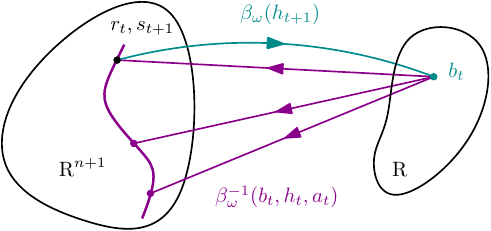}
	\vspace{-0.3cm}
	\caption{Sketch of transformation of variables $\beta_\omega$.}
	\label{fig:transformation}
	\vspace{-0.5cm}
\end{figure}

 By making a Lipschitz assumption, we answer the final part of Question I to find an exact relationship between a model-based and the equivalent model-free approach:
 \begin{assumption}\label{ass:regularity}
 	In addition to \cref{ass:measurable}, assume $Q_\omega(s,a)$ is Lipschitz in $s$ for all $\omega\in\Omega$,  $a\in\mathcal{A}$ and  $\lvert J_\beta(r_t,s_{t+1})\rvert\coloneqq\lVert \nabla_{r_t,s_{t+1}} \beta_\omega(h_{t+1})\rVert_2>0$ a.e.
 \end{assumption} 
  To determine the relationship between model-free and model-based approaches, we must study the \emph{pre-image} of the bootstrap function defined in \cref{sec:model-free-formulation}:
\begin{align}
	\beta^{-1}_\omega(b_t,h_t,a_t) \coloneqq \{r_t,s_{t+1}\vert \beta_\omega(h_{t+1}) =b_t \}.
\end{align}
Intuitively, $\beta^{-1}_\omega(b_t,h_t,a_t)$ returns all of the reward-next state pairs that produce an equivalent $b_t$, thereby mapping from Bellman operator space (whose uncertainty is characterised by model-free methods) to reward-state space (whose uncertainty is characterised by model-based methods). We sketch this mapping \cref{fig:transformation}. 
We now derive the exact Bellman distribution model used by model-free approaches starting from a given reward-state transition model as specified by a model-based approach:
\begin{corollary} \label{proof:density_derivation} Let \cref{ass:regularity}  hold. If there exists a $\omega^\star$ satisfying $\mathcal{B}^\star[Q_{\omega^\star}](h_t,a_t)=Q_{\omega^\star}(h_t,a_t)$, then the Bayes-optimal policy under a parametric reward-state transition model with density $p(r_t,s_{t+1}\vert s_t,a_t,\phi)$ and prior $P_\Phi$ is equivalent to a Bayes-optimal policy using the same prior under a optimal Bellman model with density:
	\begin{align}
		&p(b_t\vert h_t,a_t,\phi;\omega)\\
		&= \mathbb{E}_{r_t,s_{t+1}\sim P_{R,S}( s_t,a_t,\phi)}\left[\frac{  \delta(r_t,s_{t+1}\in \beta^{-1}_\omega(b_t,h_t,a_t))}{\lvert J_\beta(r_t,s_{t+1})\rvert}\right].
	\end{align}
	\vspace{-0.9cm}
\end{corollary}
\cref{proof:density_derivation}  relies on the coarea formula \citep[Theorem 3.2.12]{federer1969geometric}, which generalises the change of variables formula to non-injective mappings between dimensions. Like in SurVAE flows \citep{Nielsen20} the change of variables for non-injective mappings must be stochastic in the reverse direction $b_t\rightarrow r_t,s_{t+1}$ to account for the fact that several variables can map to a single $b_t$. $P_{R,S}(b_t,h_t,a_t;\omega)$ is thus a mapping to the set of probability distributions over the pre-image  $\beta_\omega^{-1}(b_t,h_t,a_t)$ (see \cref{fig:transformation} for a sketch). Due to \cref{ass:regularity},  $\beta_\omega^{-1}(b_t,h_t,a_t)$ is an $n$-dimensional manifold in $\mathbb{R}^{n+1}$ space, for example a $1$-dimensional curved line in $\mathbb{R}^2$ as shown in our sketch.

 Accurately representing the aleatoric uncertainty through the model $P_B( h_t,a_t;\omega)$ is the focus of distributional RL \citep{Bellemare17} and has been ignored by the model-free BRL community. As discussed in \cref{sec:related}, most existing parametric model-free BRL approaches have focused on representing the epistemic uncertainty in the posterior under a parametric Gaussian model \citep{Osband18}. One notable exception is model-based $Q$-variance estimation \citep{Luis23}. However, this approach is derived from BootDQN+Prior \citep{Osband18} which, as we prove in \cref{sec:qbrl}, forfeits Bayes-optimality.  
 
  Using \cref{proof:density_derivation}, we investigate whether a Gaussian distribution is appropriate.  If the function $\beta_\omega(h_{t+1})$ is bijective with inverse $r_t,s_{t+1}=\beta_\omega^{-1}(b_t,h_t,a_t)$ then the density is: 
  \begin{align}
  p(b_t\vert h_t,a_t,\phi;\omega)=\frac{ p(\beta_\omega^{-1}(b_t,h_t,a_t)\vert s_t,a_t, \phi)}{\lvert J_\beta(\beta_\omega^{-1}(b_t,h_t,a_t))\rvert}.\label{eq:gaussian_transform}
  \end{align}
  \cref{eq:gaussian_transform} demonstrates that even in simple MDPs with bijective transformation of variables, the equivalent space of reward-state transition distributions cannot be modelled well by Gaussian models over Bellman operators as the density $	p(b_t\vert h_t,a_t,\phi;\omega)$ can be arbitrarily far from Gaussian depending upon the choice of function approximator. This issue has been investigated empirically when modelling uncertainty in $Q$-functions \citep{Janz19}, where improving the representative capacity of a Gaussian model using successor features reduces the learning time from $\mathcal{O}(L^3)$ to $\mathcal{O}(L^{2.5})$ in the $L$-episode length chain task \citep{Osband18} under a posterior sampling exploration regime.
 This issue is particularly pertinent if we are concerned with finding polices that approach Bayes-optimality. Epistemic uncertainty estimates are rendered useless if the space of MDPs that the agent is uncertain over does not reflect the agent's environment. The key insight is that accurately representing both aleatoric and epistemic uncertainty is crucial for learning Bayesian policies with successful exploration strategies as epistemic uncertainty cannot be considered in isolation from aleatoric uncertainty. 
\vspace{-0.2cm}
\subsection{Benefits of Model-Free over Model-Based BRL}	\vspace{-0.1cm}
As discussed in \cref{sec:introduction}, obtaining exact Bayes-optimal policies is hopelessly intractable for all but the simplest domains. Models with many parameters $\Phi\subseteq \mathbb{R}^d$ exacerbate this problem as inferring a posterior and carrying out posterior marginalisation becomes exponentially more computationally complex in $d$ \citep{Bellman61}. With this in mind, we answer Question II by showing that model-free approaches need fewer parameters than equivalent model-based approaches. 

As many real-world problems of interest have high dimensional state spaces $\mathcal{S}\subseteq \mathbb{R}^n$, representing the state transition distribution in model-based approaches accurately requires a model $P_S( s,a, \phi)$ with a large number of parameters. By contrast, $b_t\in \mathbb{R}$ and hence representing the distribution $P_B^\star(h_t,a_t;\omega)$ accurately requires a model $P_B(h_t,a_t,\phi;\omega)$ with fewer parameters. Formally, the sample efficiency for density estimation of conditional distributions scales poorly with increasing dimensionality \citep{,Grunewalder12}: \citet{Wasserman06} show that when using a nonparametric frequentist kernel approach to density estimation, even with optimal bandwidth, the mean squared error scales as $\mathcal{O}(N^{\frac{-4}{n+4}})$ where $N$ is the number of samples from the true density. A mean squared error of less than 0.1 when the target density is a multivariate Gaussian of dimension 10 requires 842,000 samples compared to 19 for a two-dimensional problem. 

 Empirical results verify theoretical analysis that density estimation using normalising flows scales poorly with increasing dimension of the target distribution \citep{Papamakarios21}. From a Bayesian perspective, we also expect the posterior to concentrate at a slower rate with increasing dimensionality as the agent requires more data to decrease its uncertainty in the transition model parameters. As we discuss in \cref{app:model-based_approaches}, tractable model-based approaches based on VariBAD \citep{Zintgraf19} circumvent this issue by only inferring a posterior over a small subset of  model parameters $m\subset \phi$ at the expense of Bayes-optimality.
 
An additional benefit of modelling uncertainty in the Bellman operator is that, as we sketch in \cref{fig:transformation}, the transformation of variables $\beta_\omega(h_{t+1})$ is surjective. This reflects the fact that several reward-state transition distributions can yield the same optimal Bellman operator. As such, a single hypothesis over Bellman operator models may account for several equivalent hypotheses over reward-state transition models. As we proved in \cref{proof:model-free_model-based} that it does not matter which variable we model uncertainty in, there may be unnecessary information in reward-state transition models that becomes redundant when finding a Bayes-optimal policy. 

We now answer Question III by analysing the two families of existing model-free approaches, myopic BRL and QBRL, showing that both methods make approximations that prevent them from learning Bayes-optimal policies. \vspace{-0.2cm}
 \subsection{Myopic BRL}\vspace{-0.1cm}
 \label{sec:myopic}
The most common approximation to exact BRL, whether intentional or not, is to solve a variation of the true BAMDP where the epistemic uncertainty update is \emph{myopic}. Here the distribution: $P_{R,S}(h_t,s_{t'},a_{t'})=\mathbb{E}_{\phi \sim P_\Phi(h_t)}\left[ P_{R,S}(s_{t'},a_{t'},\phi)\right]$ is used to characterise the epistemic uncertainty over all future timesteps $t'\ge t$ and does not account for how the posterior evolves after each transition through the Bayesian Bellman operator. The corresponding distribution over a trajectory  $h_{t:t'}$ from timestep $t$ to $t'$ having observed $h_t$ is: $p^\pi_\textrm{Myopic}(h_{t:t'}\vert h_t)=\prod_{i=t}^{t'-1}\pi(a_{i}\vert s_{i}, h_t)\cdot p_{R,S}(r_i,s_{i+1}\vert s_i,a_i, h_t)$. Here, only $P_\Phi(h_t)$ is used to marginalise over uncertainty at each timestep $t'\ge t$ and information in $h_{t:t'}$ is not used to update the posterior.

Several existing model-free approaches \citep{Kuss03,Engel05,Gal16a,Fortunato18,Lipton18,Touati19}  naively introduce a $Q$-function approximator $Q_\omega:\mathcal{S}\times\mathcal{A}\rightarrow \mathbb{R}$ whose parameters minimise the mean-squared Bayesian Bellman error: $\omega^\star\in \argmin_{\omega\in\Omega} \lVert Q_\omega(s,a)- 	\mathcal{B}_\textrm{Myopic}^\star[Q_\omega](h_t,s,a)\rVert_\rho^2$ where  $	\mathcal{B}_\textrm{Myopic}^\star[Q_\omega]$ is the myopic Bellman operator: $\mathcal{B}_\textrm{Myopic}^\star[Q_\omega](h_t,s_{t'},a_{t'})=\mathbb{E}_{r_{t'},s_{t'+1}\sim P_{R,S}( h_t,s_{t'},a_{t'})}\left[ r_{t'}+\gamma \max_{a'} Q_\omega(s_{t'+1},a') \right]$.

This Bellman operator does not propagate epistemic uncertainty across timesteps; hence myopic BRL is optimal with respect to myopic beliefs and is not Bayes-optimal. Recent notable exceptions are BootDQN+Prior \citep{Osband18,Osband19}, its actor-critic analogue BBAC \citep{Fellows21} and model-based $Q$-variance estimation \citep{Luis23}; whilst these approaches satisfy an uncertainty Bellman equation \citep{odonoghue18a}, they can still be far from Bayes optimal, as we now show. 
\vspace{-0.2cm}
 \subsection{QBRL}
\label{sec:qbrl}
\vspace{-0.1cm}
 BootDQN+Prior and BBAC assume that Bayesian value functions can be decomposed using an expectation over a contextual value function using the posterior. For example, for the optimal $Q$-function: $Q_\textrm{QBRL}^\star(h_t,a_t)=\mathbb{E}_{\phi\sim P_\Phi(h_t)}\left[Q^\star(s_t,a_t,\phi)\right]$ where $Q^\star(s_t,a_t,\phi)$ is the contextual optimal $Q$-function introduced in \cref{sec:contextual_rl}. An optimal QBRL policy can be constructed using $a^\star\in \argmax_{a'}Q^\star_\textrm{QBRL}(h_t,a')$ and we denote the set of all optimal  QBRL policies as $\Pi_\textrm{QBRL}$.  This approach is analogous to the QMDP approximation used to solve POMDPs \citep{Littman95}, which amounts to assuming that uncertainty in the agent's current belief over $\phi$ disappears after the next action. We refer to these methods as QBRL. 
 
In QBRL, practical algorithms introduce a function approximator $Q_\omega(s_t,a_t,\phi)$ to estimate the contextual $Q$-function. A dataset $\mathcal{D}(h_t)$ of bootstrapped samples $b_t\coloneqq r_t +\gamma\max_{a'} Q_\omega(s_{t+1},a_{t+1},\phi)$ and a predictive distribution over $b_t$ is inferred following the formulation in \cref{sec:model-free-formulation}. Whilst both QBRL methods use posterior sampling to avoid costly posterior marginalisations, ignoring this approximation, it has not been established whether QBRL methods have the potential to learn Bayes-optimal policies, that is whether $\Pi_\textrm{QBRL}^\star=\Pi_\textrm{Bayes}^\star$. To answer this question, we first define the set of QBRL policies: $\Pi_\textrm{QBRL}\coloneqq\left\{\mathbb{E}_{\phi\sim  P_\Phi(\mathcal{H}_t)}\left[ \pi(\cdot,\phi)\right]\vert \pi \in \Pi_\Phi\right\}\subset \Pi_\mathcal{H}$. Using the following theorem, we show that QBRL approaches maximise the Bayesian RL objective but the set of policies they optimise over is restricted to $\Pi_\textrm{QBRL}$:
\begin{theorem} \label{proof:contextual_Bayes} Under \cref{ass:regularity},\vspace{-0.2cm}
	\begin{align}
	\Pi^\star_{\textnormal{\textrm{QBRL}}}&=\argmax_{\pi\in\Pi_\textrm{QBRL}} J^\pi_\textrm{Bayes},\\
	&=\left\{\mathbb{E}_{\phi\sim P_\Phi(h_t)} \left[ \pi^\star(\cdot,\phi )\right]\vert \pi^\star(\cdot,\phi)\in \Pi^\star_\Phi\right\}.
	\end{align}\vspace{-0.8cm}
\end{theorem}
\cref{proof:contextual_Bayes} proves that the set of contextual optimal policies $\Pi^\star_{\textnormal{\textrm{Contextual}}}$ can only be formed from a mixture of optimal policies conditioned on specific MDPs using the posterior. We thus prove that QBRL optimal policies can be arbitrarily Bayes-suboptimal in \cref{proof:counterexample_MDP}, using the tiger problem \citep{Kaelbling98} as a counterexample:
\begin{corollary} \label{proof:counterexample_MDP} There exist MDPs with priors such that $\Pi_\textrm{QMDP}^\star \cap \Pi_\textrm{Bayes}^\star   =\varnothing$.  \vspace{-0.2cm}
\end{corollary}
Finally, we note that a recent algorithm EVE \citep{Schmitt23} uses a combination of both myopic and QBRL approximation together.

\vspace{-0.2cm}
\section{Bayesian Exploration Network (\textsc{ben})}\vspace{-0.2cm}
\label{sec:Bayesian_exploration_networks}
Using our formulation in \cref{sec:myopic}, we develop a model-free BRL algorithm  capable of learning Bayes-optimal policies. As we are taking a value-based approach in this paper, we focus on solving the optimal Bayesian Bellman equation. We now  introduce the Bayesian Exploration network (\textsc{ben}), which is comprised of three individual networks: a $Q$-network to reduce the dimensionality of inputs to a one-dimensional variable and then two normalising flows to characterise both the aleatoric and epistemic uncertainty over that variable as it passes through the Bellman operator. 
\vspace{-0.6cm}
\subsection{Network Specification}\vspace{-0.1cm}
\label{sec:network_spec}
\paragraph{Recurrent $Q$-Network} For our choice of function approximator,  we encode history using a recurrent neural network (RNN). Similar approximators are used in POMDP solvers \citep{Hausknecht15, schlegel2023}. The $Q$-function approximator is a mapping from history-action pairs, allowing uncertainty to propagate properly through the Bayesian Bellman equation. By contrast, encoding of history is missing from state-of-the-art model-free approaches based on QBRL as these function approximators condition on a context variable instead. We denote the output of the function approximator at time $t$ as $q_t=Q_\omega(h_t,a_t)$.\vspace{-0.3cm}
\paragraph{Aleatoric Network}
We now specify our proposed model $P_B(h_t,a_t,\phi;\omega)$. Recall from \cref{sec:aleatoric} that the performance of model-free BRL depends on the capacity for representing aleatoric uncertainty. Whilst the model in \cref{proof:density_derivation} could be used in principle, there are two practical issues: firstly, we cannot assume knowledge of the preimage $b^{-1}_\omega(b_t,h_t,a_t)$ for arbitrary function approximators; and, secondly, the transformation $b_\omega(h_{t+1})$ is a surjective mapping  $\mathbb{R}^{n+1}\rightarrow\mathbb{R}$, meaning we would not be taking advantage of the projection down to a lower dimensional space by first specifying a model $P_{R,S}(s_t,a_t,\phi)$. 

 Instead, we specify $P_B(h_t,a_t,\phi;\omega)$ using a normalising flow for density estimation \citep{Rezende15,Kobyzev19}, making a transformation of variables $b_t=B(z_\textrm{al},q_t,\phi)$ where $z_\textrm{al}\in\mathbb{R}$ is a base variable with a zero-mean, unit variance Gaussian $P_\textrm{al}$. Sampling $b_t\sim P_B(h_t,a_t,\phi;\omega)$ is equivalent to sampling $z_t\sim P_\textrm{al}$ and then applying the transformation $b_t=B(z_\textrm{al},q_t,\phi)$ . Details can be found in \cref{app:aleatoric}. We refer to this density estimation flow as the \emph{aleatoric} network as it characterises the aleatoric uncertainty in the Bellman operator. Unlike in model-based approaches where the hypothesis space must be specified a-priori, in \textsc{ben} the hypothesis space is determined by the representability of the aleatoric network, which can be tuned to the specific set of problems. Under mild regularity assumptions \citep{Huang18}, an autoregressive flow as a choice for the aleatoric network can represent any target distribution $P_B(h_t,a_t,\phi;\omega)$ to arbitrary accuracy given sufficient data \citep{Kobyzev19}. 

A key advantage of our approach is that we have preprocessed the input to our aleatoric network through $q_t=Q_\omega(h_t,a_t)$ to extract features that reduce the dimensionality of the state-action space.  This architecture hard-codes the prior information that a Bellman operator is a functional of the $Q$-function approximator, meaning that we only need to characterise aleatoric uncertainty in a lower dimensional input $q_t$. Unlike in VariBAD, we do not need to introduce frequentist heuristics to learn function approximator parameters $\omega$. Instead these are learnt automatically by solving the optimal Bayesian Bellman equation, which we detail in \cref{sec:msbbe}.  \vspace{-0.2cm}
\paragraph{Epistemic Network} 
  Inferring the posterior and carrying out marginalisation exactly is intractable for all but the simplest aleatoric networks, which would not have sufficient capacity to represent a complex target  distribution $P_B(h_t,a_t,\phi;\omega)$. Instead we use variational inference via normalising flows to learn a tractable approximation $P_\psi$ parametrised by $\psi\in \Psi$ which we learn by minimising the KL-divergence between the two distributions $\kl{P_\psi}{P_\Phi(\mathcal{D}_\omega(h_t))}$. This is equivalent to minimising the tractable evidence lower bound $\textrm{ELBO}(\psi; h,\omega)$. We provide details in  \cref{app:epistemic}. We refer to this flow as the epistemic network as it characterises the epistemic uncertainty in $\phi$. To our knowledge, \textsc{ben} is the first time flows have been used for combined density estimation and variational inference. 
\vspace{-0.2cm}
\subsection{Mean Squared  Bayesian Bellman Error (MSBBE)}\vspace{-0.1cm}
\label{sec:msbbe}
Finally, we learn a parametrisation $\omega^\star$ that satisfies the optimal Bayesian Bellman equation for our $Q$-function approximator. For \textsc{ben}, this is equivalent to minimising the Mean Squared  Bayesian Bellman Error (MSBBE) between the predictive optimal Bellman operator $B^+[Q_\omega]$ and $Q_\omega$:
\begin{align}
	\textrm{MSBBE}(\omega;h_t,\psi)\coloneqq\left\lVert  B^+[Q_\omega](h_t,a_t) -Q_{\omega}(h_t,a_t)   \right\rVert_{\rho}^2,
\end{align}
 where $\rho$ is an arbitrary sampling distribution with support over $\mathcal{A}$. Given sufficient compute, at each timestep $t$ it is possible in principle to solve the nested optimisation problem for \textsc{ben}: $\omega^\star\in \argmin_{\omega\in\Omega} 	\textrm{MSBBE}(\omega;h_t,\psi^\star(\omega)), \textrm{such that}\  \psi^\star(h_t,\omega)\in \argmin_{\psi\in\Psi}\textrm{ELBO}(\psi; h,\omega)$. Nested optimisation problems are commonplace in model-free RL and can be solved using two-timescale stochastic approximation: we update the epistemic network parameters $\psi$ using gradient descent on an asymptotically faster timescale than the function approximator parameters $\omega$ to ensure convergence to a fixed point \citep{Borkar08,Heusel17,Fellows21}, with $\omega$ playing a similar role as target network parameters used to stabilise TD. 
	\begin{algorithm}\vspace{-0.1cm}
	\caption{$\textsc{ApproxBRL}(P_\Phi,\mathcal{M}(\phi))$}
	\label{alg:approxBRL}
	\begin{algorithmic}
		\STATE Initialise $\omega,\psi$,  $h = s$
		\STATE Sample initial state $s \sim P_0$
		\STATE Take $N_\textrm{Pretrain}$ SGD Steps on $\textrm{MSBBE}(\omega)$
		\WHILE{posterior \textbf{not} converged}
		\STATE Take action $a \in \argmax_{a'}Q_{\omega}(h,a') $
		\STATE Observe reward $r \sim P_R(s,a,\phi^\star)$
		\STATE Transition to new state $s \sim P_S(s,a,\phi^\star)$
		\STATE $h \leftarrow \{h,a,r,s \}$
		\FOR{$N_\textrm{Update}$ Steps: }
		\STATE Take $N_\textrm{Posterior}$ SGD steps on $\textrm{ELBO}(\psi; h,\omega)$
		\STATE Take a SGD step on $\textrm{MSBBE}(\omega;h,\psi)$
		\ENDFOR
		\ENDWHILE
	\end{algorithmic}
\end{algorithm} 
\vspace{-0.3cm}
Solving the MSBBE exactly for every observable history $h_t$ is computationally intractable. Instead, we propose partial minimisation of our objectives as outlined in \cref{alg:approxBRL}: after observing a new tuple $\{a,r,s\}$, the agent takes $N_\textrm{Update}$ MSBBE update steps using the new data. This is equivalent to partially minimising the empirical expectation $\mathbb{E}_{h\sim h_t}	\left[\textrm{MSBBE}(\omega;h,\psi^\star(\omega))\right]$, where each $h\sim h_t$ is a sequence drawn from the observed history, analogously to how state-action pairs are drawn from the replay buffer in DQN \citep{mnih16}. To ensure a separation of timescales between parameter updates, the agent carries out $N_\textrm{Posterior}$ steps of SGD on the ELBO for every MSBBE update. Our algorithmic is shown in \cref{alg:approxBRL}.

Additionally, we exploit the fact that the MSBBE can be minimised prior to learning using samples of state-action pairs and so carry out $N_\textrm{Pretrain}$ pretraining steps of SGD on the loss using the prior in place of the approximate posterior. If no prior knowledge exists, then the agent can be deployed. If there exists additional domain-specific knowledge, such as transitions shared across all MDPs or demonstrations at key goal states, this can also be used to train the agent using the model-based form of the Bellman operator. Full algorithmic details can be found in \cref{app:nework_training}.
 \vspace{-0.3cm}
\section{Experiments} 
\begin{figure}
 	\vspace{-0.1cm}
 	\centering
 	\includegraphics[width=1.04\columnwidth]{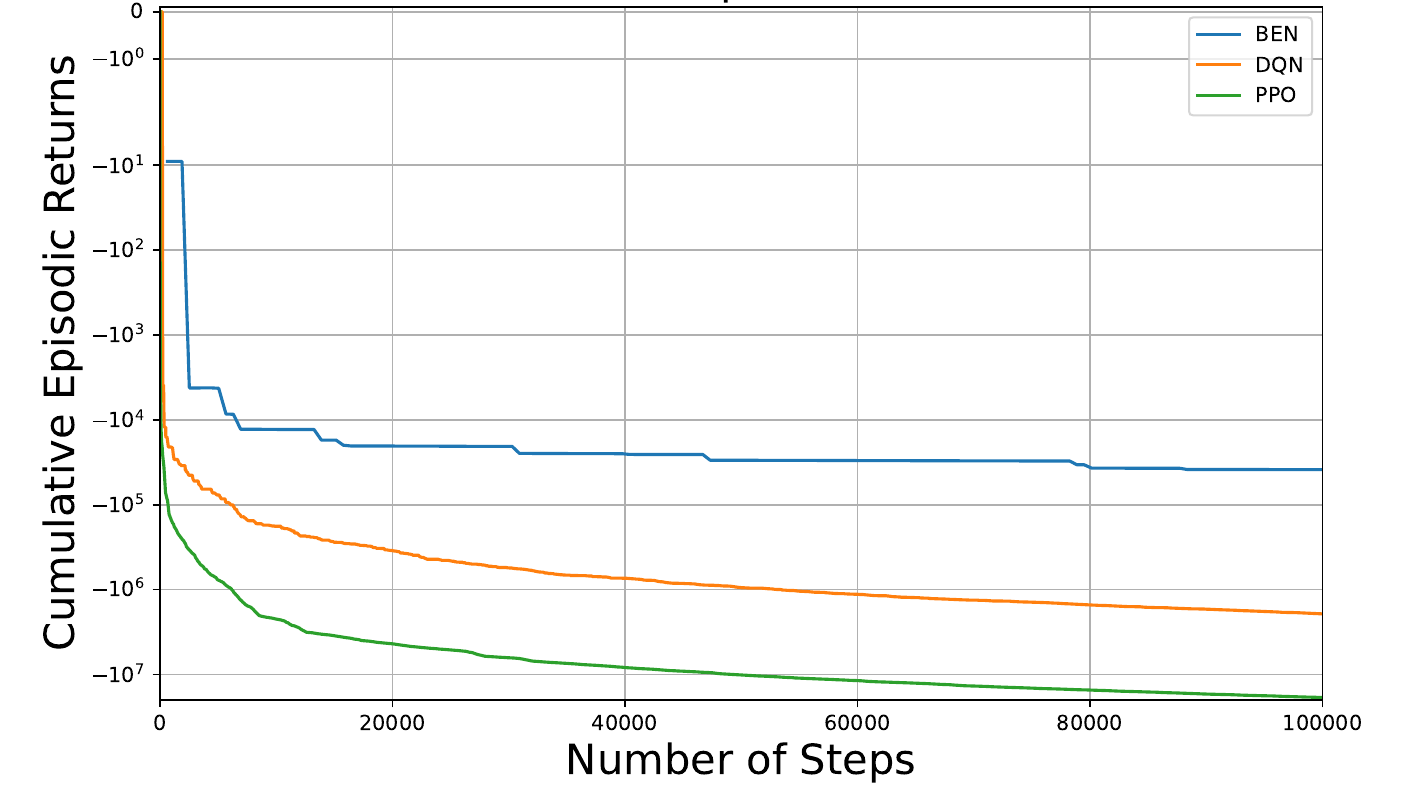}
 	\vspace{-0.7cm}
 	\caption{Evaluation in search and rescue episodic problem with no prior knowledge, showing cumulative return of \textsc{ben} vs RNN PPO and BootDQN+prior.} 
 	\label{fig:ben_v_sota}
 	\vspace{-0.4cm}
 \end{figure}

\begin{figure}[h]
 	\vspace{-0.2cm}
 	\centering
 	\includegraphics[width=0.7\columnwidth]{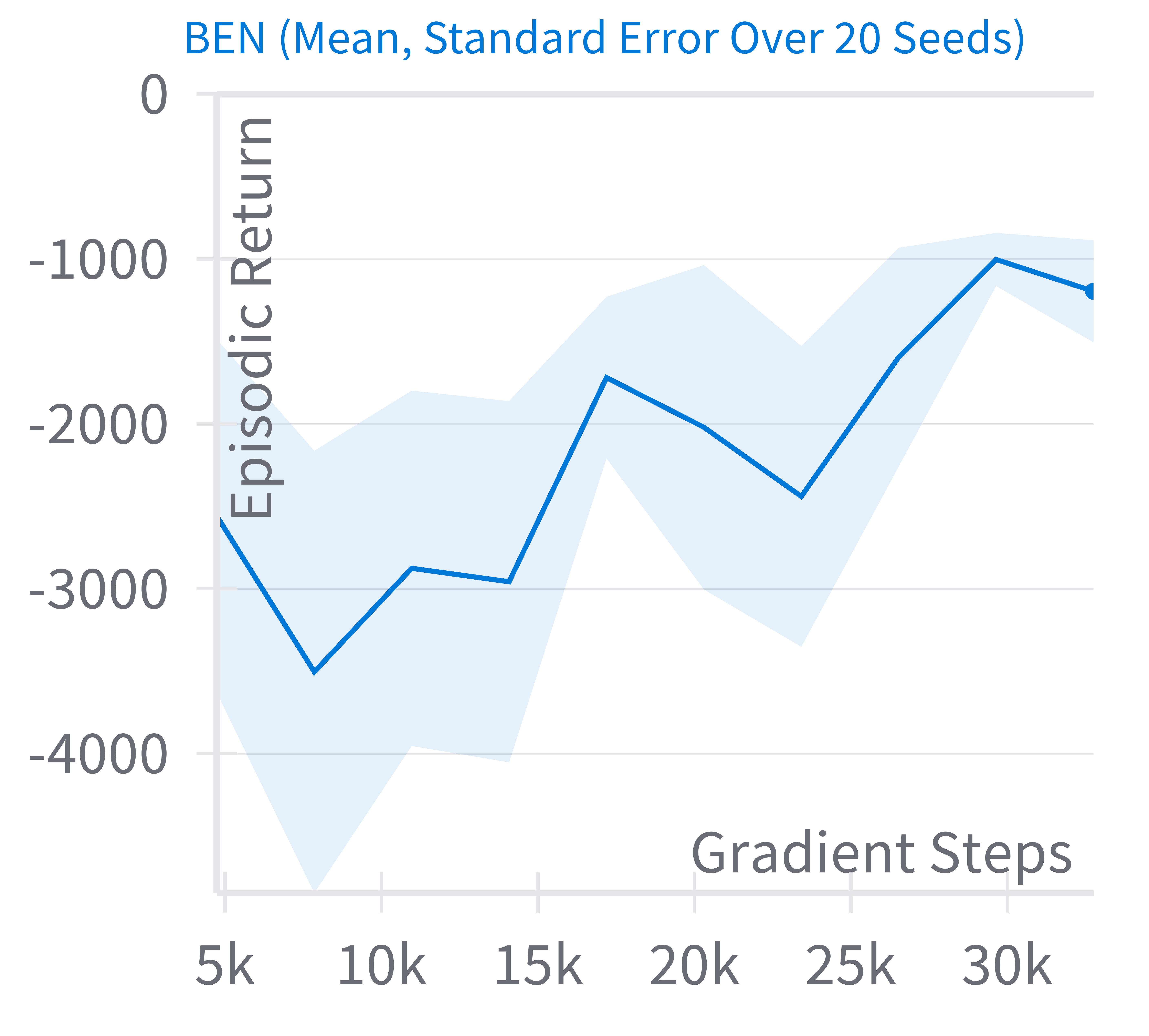}
 	\vspace{-0.4cm}
 	\caption{Evaluation in search and rescue episodic problem with weak prior knowledge, showing return of \textsc{ben} after each episode.} 
 	\label{fig:search_rescue_results}
 	\vspace{-0.4cm}
 \end{figure}

\begin{figure}[H]
	\vspace{-0.3cm}
	\centering
	\includegraphics[width=0.7\columnwidth]{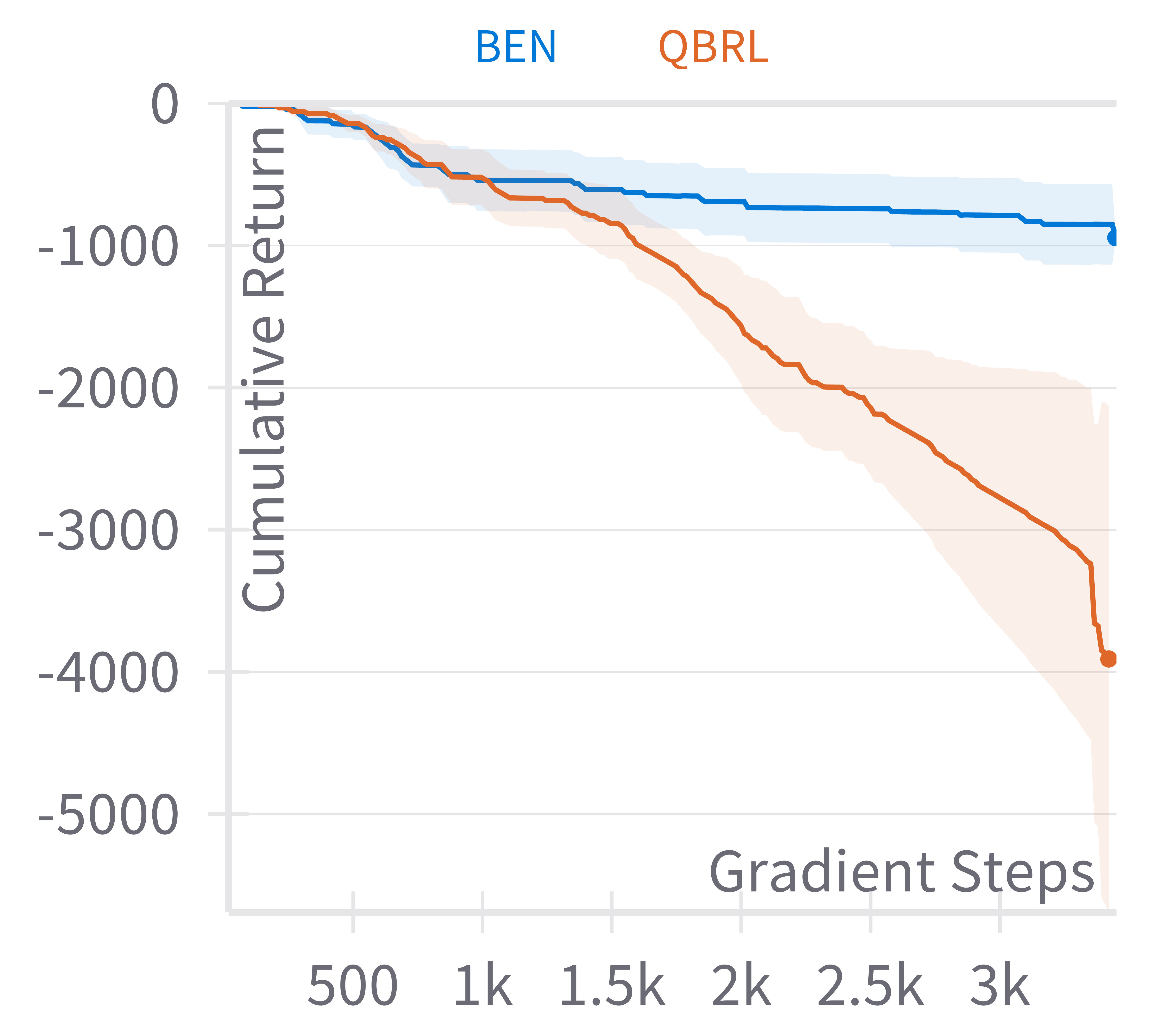}
	\vspace{-0.3cm}
	\caption{Evaluation in zero-shot search and rescue showing cumulative return for \textsc{ben} vs.\ QBRL methods.}
	\vspace{-0.3cm}
	\label{fig:search_one_results}
\end{figure}

\vspace{-0.1cm}
\label{sec:experiments}
We introduce a novel search and rescue gridworld MDP as a more challenging, higher-dimensional extension to the tiger problem (which we show \textsc{ben} can solve in \cref{app:experiments}). An agent is tasked with rescuing $N_\textrm{victims}$ victims from a dangerous situation whilst avoiding any one of $N_\textrm{hazards}$ hazards. Details can be found in \cref{app:search_rescue_problem}. We evaluate \textsc{ben} using a $7\times 7$ grid size with 8 hazards and 4 victims.
 \vspace{-0.3cm}
 \paragraph{Episodic Setting, Tabula Rasa}  In the episodic setting, the environment is reset after 245 timesteps (5 times the grid size to allow time to visit the edges of larger grids), and a new environment is uniformly sampled from the space of MDPs. After resetting, the epistemic parameters $\psi$ are also reset, representing the belief in the new MDP returning to the prior. $Q$-network parameters $\omega$ are retained so the agent can exploit information that is shared across MDPs.  We initialise the agent with a zero-mean Gaussian prior of diagonal variance equal to $0.1$. We assume no prior environment knowledge of any kind in this experiment. We compare to PPO using an RNN to encode history as a strong frequentist baseline. We also compare to BootDQN+prior, which is the state-of-the-art Bayesian model-free method. Results are shown in \cref{fig:ben_v_sota} where we plot the cumulative return for the three methods.
 Whilst BootDQN+prior can outperform RNN PPO due to its more sophisticated deep exploration, it cannot successfully learn to listen to solve the task. This confirms our key theoretical analysis that methods based on QBRL learn contextual policies that can be arbitrarily suboptimal (\cref{proof:counterexample_MDP}). Unlike BEN, BootDQN+prior cannot be used to approximate Bayes-optimal policies. 
\vspace{-0.3cm}
  \paragraph{Episodic Setting, Weak Prior}
 We repeat the experiment in the episodic setting, this time showing the agent examples of deterministic movement and the average reward for opening a door at random as a weak prior. The results for our implementation are shown in \cref{fig:search_rescue_results}. We plot the return at the end of each 245 timestep episode. As expected, \textsc{ben} can solve this challenging problem, exploring in initial episodes to learn about how the listening states correlate to victim and hazard positions, then exploiting this knowledge in later episodes, finding all victims immediately. Our results demonstrate that \textsc{ben} can scale to higher dimensional domains without forfeiting BRL's strong theoretical properties through approximation. 
 \vspace{-0.2cm}
 \paragraph{Zero-shot Setting, Strong Prior}
 In this setting, our goal is to investigate how effectively \textsc{ben} can exploit strong prior knowledge to solve the search and rescue environment in a single episode. We prior-train \textsc{ben} using simulations in related (but not identical) environments drawn from a uniform prior, showing the agent the effect of listening. This experiment is designed to mimic a real-life application where simulations can be provided by demonstrators in a generic training setting, followed by deployment in a novel environment where the robot has only one chance to succeed. Details can be found in \cref{app:search_and_rescue_prior}. To investigate effect of history dependency on the solution, we also compare to a variant of BEN using a $Q$-function that only conditions on state-actions, representing an ideal QBRL solution. We plot the cumulative return as a function of number of gradient steps over the course of the episode in \cref{fig:search_one_results} for both \textsc{ben} and the QBRL variant. Our results demonstrate that by exploiting prior knowledge, \textsc{ben} can successfully rescue all victims and avoid all hazards, even when encountering a novel environment. In contrast, the oracle policy for existing state-of-the-art model-free methods, which learn a QBRL Bayes policy, cannot solve this problem because, as our analysis in  \cref{sec:qbrl} shows, $\Pi^\star_\textrm{QBRL}$ is limited to mixtures of optimal policies conditioned on $\phi$, causing contextual agents to repeatedly hit hazards. This challenging setting showcases the high sample efficiency with low computational complexity of \textsc{ben}.
 In addition, we perform two ablations in \cref{sec:ablations}. First, we demonstrate that performance depends on the capacity of the aleatoric network, verifying our claim in \cref{sec:network_spec} that there is a balancing act specifying a rich enough hypothesis space to represent the true model accurately but that is small enough to generalise effectively. Secondly, we investigate how pre-training affects returns. As we decrease the number of prior pre-training MSBBE minimisation steps, we see that performance degrades in the zero-shot settling, as expected. Moreover, this ablation shows that relatively few pre-training steps are needed to achieve impressive performance once the agent is deployed in an unknown MDP.

\vspace{-0.3cm}\section{Conclusions}\vspace{-0.2cm}
We carried out theoretical analyses of model-free BRL and formulated the first model-free approach that can learn Bayes-optimal policies. We proved that existing model-free approaches for BRL are limited to optimising over a set of QBRL policies or optimise a myopic approximation of the true BRL objective. In both cases, the corresponding optimal policies can be arbitrarily Bayes-suboptimal. We introduced \textsc{ben}, a model-free BRL algorithm that can successfully learn true Bayes-optimal policies. Our experimental evaluation confirms our analysis, demonstrating that \textsc{ben} can behave near Bayes-optimally even under partial minimisation, paving the way for a new generation of model-free BRL approaches with the desirable theoretical properties of model-based approaches.

 \clearpage
  \section*{Acknowledgements}
  Mattie Fellows is funded by a generous grant from Waymo. Christian Schroeder de Witt is funded by UKRI
grant: Turing AI Fellowship EP/W002981/1, Armasuisse Science+Technology, and an EPSRC IAA Impact Fund award. The experiments were made possible by a generous equipment grant from NVIDIA. 
 \section*{Impact Statement}
 \label{sec:impact}
 This paper presents work whose goal is to advance the field of Bayesian reinforcement learning. Our primary contribution is theoretical and there are no specific dangers of our work. Any advancement in RL should be seen in the context of general advancements to machine learning. Whilst machine learning has the potential to develop useful tools to benefit humanity, it must be carefully integrated into underlying political and social systems to avoid negative consequences for people living within them. A discussion of this complex topic lies beyond the scope of this work.

\bibliography{brl}
\bibliographystyle{icml2024/icml2024}

\newpage
\appendix
\onecolumn
\section{Model-Based Approaches and their Shortcomings}
\label{app:model-based_approaches}

Whilst POMDP solvers such as $\textrm{RL}^2$ \citep{duan2017rl} or meta-learning approaches \citep{Beck23} can be naively applied to the BAMDP, solving the BAMDP exactly is hopelessly intractable for all but the simplest of problems for several reasons: firstly, unless a conjugate model is used, the posterior over model parameters cannot be evaluated analytically as the posterior normalisation constant is intractable to evaluate and conjugate models are often too simple to be of practical value in RL; secondly, even if it is possible to obtain a posterior over the model parameters, solving the BAMDP requires evaluating high dimensional integrals by marginalising over model parameters; finally, finding a Bayes-optimal policy in the MDP requires solving a planning problem at each timestep for all possible histories. Few model-based BRL algorithms scale beyond small and discrete state-action spaces, even under approximation \citep{Asmuth11,Guez13}. 

One notable exception is the VariBAD framework \citep{Zintgraf19}, and subsequent related approaches \citep{Lee21,zintgraf21a}, which avoids the problem of intractability by being Bayesian over a small subset of model parameters. These methods sacrifice Bayes-optimally, relying on a frequentist heuristic to learn the non-Bayesian parameters; here, parametric models of the reward and state and transition distributions are introduced: $P_\theta(r\vert s,a,m)$ and $P_\theta(r\vert s,a,m)$ which are parametrised by $\theta\in\Theta$ and condition on the random variable $m$. A posterior over $m$ is then inferred and used to obtain a marginal likelihood over trajectories:
\begin{align}
	p(\tau\vert \theta,\mathcal{D})=\mathbb{E}_{m\sim p(m\vert \mathcal{D})}\left[ p_\theta(\tau\vert m) \right],
\end{align}
which is optimised for $\theta$. In this way, VariBAD is a partially Bayesian approach, inferring a posterior over $m$, but not parameters $\theta$.  The dimensionality of $m$ can be kept relatively small to ensure tractability. As the VariBAD optimal policy is not obtained by marginalising over a space of MDPs nor is the uncertainty accounted for in all of the model's parameters, it is not Bayes-optimal in the limit of approximation.

Another issue with model-based approaches, especially those trained in a meta-learning context such as $\textrm{RL}^2$, is the assumption that the agent has access to a generative hypothesis space where it is possible to sample MDPs from a prior and collect rollouts of transitions from each sampled MDP. In real-world settings, knowing the exact hypothesis space is not a feasible assumption as it is generally not possible to specify transition dynamics a priori for all environments the agent can encounter. 

\section{Proofs}
\label{app:proofs}
\renewcommand{\thesection}{\arabic{section}}
\setcounter{section}{3}
\setcounter{theorem}{1}
\begin{theorem}\label{proof_app:model_free_equivalence}
	Let \cref{ass:measurable} hold, then $B^+[Q_\omega](h_t,a_t )=B^\star[Q_\omega](h_t,a_t )$.
	\begin{proof}
		We start by analysing the transformation of variables $b_t= \beta_\omega(h_{t+1})\coloneqq r_t+\gamma \max_{a'} Q_{\omega}(h_{t+1},a')$. To ease notation, for a variable $x\in\mathbb{R}^d$ we write an integral with respect to the Lebesgue measure $\mathcal{L}^d$ as $\int f(x) d\mathcal{L}^d(x)=\int f(x) dx$. Under \cref{ass:measurable}, the mapping $\beta_\omega(\cdot,h_t,a_t):\mathbb{R}\times \mathcal{S}\rightarrow \mathbb{R}$ is also measurable, hence from the change of variables theorem under measurable mappings (see \citet[Theorem 3.6.1]{Bogachev07}):
		\begin{align}
			\mathcal{B}^\star[Q_{\omega}](h_t,a_t)&=\mathbb{E}_{r_t,s_{t+1}\sim P_{R,S}(h_t,a_t)}\left[\beta_{\omega}(h_{t+1}) \right]\\
			&=\int  \beta_{\omega}(h_{t+1}) p(r_t,s_{t+1}\vert h_t,a_t) d(r_t,s_{t+1}),\\
			&=\int b_t p(b_t\vert h_t,a_t;\omega) db_t.
		\end{align}
		Marginalising over $b_0,b_1,.. b_{t-1}$ yields:
		\begin{align}
			\mathcal{B}^\star[Q_{\omega}](h_t,a_t)&=\int b_t p(b_0,b_1,\cdots b_t\vert h_t,a_t;\omega) d(b_0,b_1,\cdots b_t),\\
			&=\int b_t p(b_t\vert b_0,b_1,\cdots b_{t-1}, h_t,a_t;\omega) dP( b_0,b_1,\cdots b_{t-1}\vert h_t;\omega)d b_t,
		\end{align}
		Now, because each $b_i$ is a deterministic transformation of variables $b_i= \beta_\omega(h_{i+1})= r_i+\gamma \max_{a'} Q_{\omega}(h_{i+1},a')$, the distribution $dP( b_0,b_1,\cdots b_{t-1}\vert h_t;\omega)= d\delta(b_0=\beta_\omega(h_1),b_1=\beta_\omega(h_2),\cdots b_{t-1}=\beta_\omega(h_t))$, hence:
		\begin{align}
				\mathcal{B}^\star[Q_{\omega}](h_t,a_t)&\int b_t p(b_t\vert b_0,b_1,\cdots b_{t-1}, h_t,a_t;\omega) d\delta(b_0=\beta_\omega(h_1),b_1=\beta_\omega(h_2),\cdots b_{t-1}=\beta_\omega(h_t))d b_t,\\
			&=\int b_t p(b_t\vert b_0=\beta_\omega(h_1)\cdots b_{t-1}=\beta_\omega(h_t), h_t,a_t;\omega) d b_t,\\
			&=\int b_t \int p(b_t\vert h_t,a_t,\phi;\omega)p(\phi\vert b_0=\beta_\omega(h_1)\cdots b_{t-1}=\beta_\omega(h_t), h_t,a_t) d\phi d b_t,\\
			&=\int b_t \int p(b_t\vert h_t,a_t,\phi;\omega)p(\phi\vert \mathcal{D}_\omega(h_t)) d\phi d b_t,\\
		 &  =B^+[Q_\omega](h_t,a_t),
		\end{align}
		as required. 
	\end{proof}
\end{theorem}  

\setcounter{theorem}{3}
\begin{corollary}  Let \cref{ass:regularity}  hold. Assume there exists a parametrisation $\omega^\star$ such that $\mathcal{B}^\star[Q_{\omega^\star}](h_t,a_t)=Q_{\omega^\star}(h_t,a_t)$. A Bayes-optimal policy under a parametric reward-state transition model with density $p(r_t,s_{t+1}\vert s_t,a_t,\phi)$ and prior $P_\Phi$ is equivalent to a Bayes-optimal policy under a optimal Bellman model with density:
	\begin{align}
		p(b_t\vert h_t,a_t,\phi;\omega)= \mathbb{E}_{r_t,s_{t+1}\sim p(r_t,s_{t+1}\vert s_t,a_t,\phi)}\left[\frac{  \delta(r_t,s_{t+1}\in \beta^{-1}_\omega(b_t,h_t,a_t))}{\lvert J_\beta(r_t,s_{t+1})\rvert}\right],\label{eq:transformed_dist}
	\end{align}
	 with the same prior.
	\begin{proof}
	We verify our result by proving $\mathcal{B}^\star[Q_{\omega}](h_t,a_t)=B^+[Q_\omega](h_t,a_t)$ for $	p(b_t\vert h_t,a_t,\phi;\omega)$ defined above, hence any agent following policy $a^\star\in \argmax_{a'} B^+[Q_{\omega^\star}](h_t,a')$ will be taking actions  $a^\star\in \argmax_{a'} Q_{\omega^\star}(h_t,a_t)=Q^\star(h_t,a_t)$ and thus following a Bayes-optimal policy. We first prove:
	\begin{align}
		\mathbb{E}_{r_t,s_{t+1}\sim P_{R,S}(s_t,a_t,\phi)} \left[ \beta_\omega(h_{t+1})\right]=\mathbb{E}_{b_t\sim P_{B}(h_t,a_t,\phi;\omega)} \left[b_t \right].
	\end{align} 
	Starting from the left hand side, as the $n+1$ dimensional Hausdorff and Lebesgue measures agree over $\mathbb{R}^{n+1}$, we write the expectation as a Lebesgue integral under the Hausdorff measure $\mathcal{H}^{n+1}$:
	\begin{align}
		\mathbb{E}_{r_t,s_{t+1}\sim P_{R,S}(s_t,a_t,\phi)} &\left[ \beta_\omega(h_{t+1})\right]\\
		&=\int_{\mathbb{R}\times \mathcal{S}}\beta_\omega(h_{t+1}) p(r_t,s_{t+1}\vert s_t,a_t, \phi) d\mathcal{H}^{n+1}(r_t,s_{t+1}),\\
		&=\int_{\mathbb{R}\times \mathcal{S}}\frac{\beta_\omega(h_{t+1}) p(r_t,s_{t+1}\vert s_t,a_t, \phi)}{\lvert J_\beta(r_t,s_{t+1})\rvert}\lvert J_\beta(r_t,s_{t+1})\rvert d\mathcal{H}^{n+1}(r_t,s_{t+1}).
	\end{align} 
		  where $\mathcal{H}^n$ is the $n$-dimensional Hausdorff measure. Because of the Lipschitz assumption under \cref{ass:regularity}, $\lvert J_\beta(r_t,s_{t+1})\rvert$ exists almost everywhere  due to Rademacher's theorem \citep[Theorem 14.25]{villani2008}, hence under \cref{ass:regularity}, we can apply the coarea formula to the transformation of variables $b_t=\beta_\omega(h_{t},a_t,r_t,s_{t+1})$ \citep[Theorem 3.2.12]{federer1969geometric}:
	\begin{align}
		&\mathbb{E}_{r_t,s_{t+1}\sim P_{R,S}(s_t,a_t,\phi)} \left[ \beta_\omega(h_{t+1})\right]\\
		&=\int_\mathbb{R}  \int_{\beta_\omega^{-1}(b_t,h_t,a_t)} b_t \frac{ p(r_t,s_{t+1}\vert s_t,a_t, \phi)}{\lvert J_\beta(r_t,s_{t+1})\rvert} d\mathcal{H}^{n}(r_t,s_{t+1}) d\mathcal{H}^{1}(b_t),\\
		&=\int_\mathbb{R}  b_t \int_{\beta_\omega^{-1}(b_t,h_t,a_t)} \frac{ p(r_t,s_{t+1}\vert s_t,a_t, \phi)}{\lvert J_\beta(r_t,s_{t+1})\rvert} \delta(r_t,s_{t+1}\in \beta^{-1}_\omega(b_t,h_t,a_t)) d\mathcal{H}^{n}(r_t,s_{t+1}) d\mathcal{H}^{1}(b_t),\\
		&=\int_\mathbb{R} b_t \mathbb{E}_{r_t,s_{t+1}\sim p(r_t,s_{t+1}\vert s_t,a_t,\phi)}\left[\frac{  \delta(r_t,s_{t+1}\in \beta^{-1}_\omega(b_t,h_t,a_t))}{\lvert J_\beta(r_t,s_{t+1})\rvert}\right]d\mathcal{H}^{1}(b_t),\\
		&=\mathbb{E}_{b_t\sim P_B(h_t,a_t,\phi;\omega)}\left[ b_t\right],
	\end{align}
	where $P_B(h_t,a_t,\phi;\omega)$ has density:
	\begin{align}
		p(b_t\vert h_t,a_t,\phi;\omega)= \mathbb{E}_{r_t,s_{t+1}\sim p(r_t,s_{t+1}\vert s_t,a_t,\phi)}\left[\frac{  \delta(r_t,s_{t+1}\in \beta^{-1}_\omega(b_t,h_t,a_t))}{\lvert J_\beta(r_t,s_{t+1})\rvert}\right],
	\end{align}
	as required.  Using \cref{proof_app:model_free_equivalence} our result follows:
	\begin{align}
		\mathcal{B}^\star[Q_{\omega}](h_t,a_t)&=\mathbb{E}_{r_t,s_{t+1}\sim P_{R,S}(h_t,a_t)}\left[\beta_{\omega}(h_{t+1}) \right]\\
		&=\mathbb{E}_{\phi\sim P_\Phi(h_t) }\left[ \mathbb{E}_{r_t,s_{t+1}\sim P_{R,S}(s_t,a_t,\phi)}\left[\beta_{\omega}(h_{t+1}) \right]\right],\\
		&=\mathbb{E}_{\phi\sim P_\Phi(h_t) }\left[ \mathbb{E}_{b_t\sim P_{B}(h_t,a_t,\phi;\omega)}\left[b_t \right]\right],\\
		&=\mathbb{E}_{\phi\sim P_\Phi(\mathcal{D}(h_t)) }\left[ \mathbb{E}_{b_t\sim P_{B}(h_t,a_t,\phi;\omega)}\left[b_t \right]\right],\\
		&=B^+[Q_{\omega}](h_t,a_t).
	\end{align}
		
	\end{proof}
\end{corollary}

\begin{theorem} \label{proof_app:bayes_value}  Under \cref{ass:regularity},
	\begin{align}
		\Pi^\star_{\textnormal{\textrm{QBRL}}}=\argmax_{\pi\in\Pi_\textrm{QBRL}} J^\pi_\textrm{Bayes}=\left\{\mathbb{E}_{\phi\sim P_\Phi(h_t)} \left[ \pi^\star(\cdot,\phi )\right]\vert \pi^\star(\cdot,\phi)\in \Pi^\star_\Phi\right\}.
	\end{align}
	\begin{proof}
		We first show $	\Pi^\star_{\textnormal{\textrm{QBRL}}}=\argmax_{\pi\in\Pi_\textrm{QBRL}} J^\pi_\textrm{Bayes}$. We use the QBRL assumption to derive the expected return:                                                   
	 \begin{align}
	 	J^{\pi^\star}_\textrm{QBRL}&=\mathbb{E}_{s_0\sim P_0} \left[V^\star_\textrm{QBRL}(s_0)\right],\\
	 	&=\mathbb{E}_{s_0\sim P_0} \left[\mathbb{E}_{\phi\sim P_\Phi}\left[V^\star(s_0,\phi)\right]\right],\\
	 	&=\mathbb{E}_{s_0\sim P_0,\phi\sim P_\Phi} \left[\mathbb{E}_{a_0\sim \pi^\star(s_0,\phi)} \left[Q^\star(s_0,a',\phi)\right]\right].
	 \end{align}
	Substituting for the definition of the optimal contextual $Q$-function:
	 \begin{align}
	 	J^{\pi^\star}_\textrm{QBRL}&=\mathbb{E}_{\phi\sim P_\Phi,s_0\sim P_0,a_0\sim \pi^\star(s_0,\phi)} \left[\mathbb{E}_{\tau_\infty\sim P_\infty^{\pi^\star}(s_0,a_0,\phi)} \left[\sum_{t=0}^\infty \gamma^t r_t\right]\right],\\
	 	&=\mathbb{E}_{\phi\sim P_\Phi} \left[\mathbb{E}_{\tau_\infty\sim P_\infty^{\pi^\star}(\phi)} \left[\sum_{t=0}^\infty\gamma^t r_t\right]\right],\\
	 	&=\mathbb{E}_{\phi\sim P_\Phi} \left[J^{\pi^\star}(\phi)\right],\\
	 	&=\mathbb{E}_{\phi\sim P_\Phi} \left[\max_{\pi\in\Pi_\Phi }J^{\pi}(\phi)\right].
	 \end{align}
	 We now prove $\Pi^\star_\Phi=\argmax_{\pi\in\Pi_\Phi} \mathbb{E}_{\phi\sim P_\Phi} \left[ J^\pi(\phi)\right] $ $P_\Phi$-almost everywhere by contradiction. We first show:
	 \begin{align}
	 	\Pi^\star_\Phi\subseteq\argmax_{\pi\in\Pi_\Phi} \mathbb{E}_{\phi\sim P_\Phi} \left[ J^\pi(\phi)\right].
	 \end{align} 
	  Assume that there exists some $\pi^\dagger(\cdot,\phi)\in \Pi^\star_\Phi\notin\argmax_{\pi\in\Phi_\Phi} \mathbb{E}_{\phi\sim P_\Phi} \left[ J^\pi(\phi)\right]  $. There then exists some set $\Phi'\subseteq\Phi$ with non-zero measure according to $P_\Phi$ such that:
	  \begin{align}
	  J^{\pi^\dagger}(\phi)<\max_{\pi}J^{\pi}(\phi),
	  \end{align}
	 for $\phi\in\Phi'$, which is a contradiction as $\pi^\dagger(\cdot,\phi)\in \argmax_{\pi}J^{\pi}(\phi)$. We are left to show:
	 	 \begin{align}
	 	\Pi^\star_\Phi\supseteq\argmax_{\pi\in\Pi_\Phi} \mathbb{E}_{\phi\sim P_\Phi} \left[ J^\pi(\phi)\right].
	 \end{align} 
	 Assume that there exists some $\pi^\dagger(\cdot,\phi)\in \argmax_{\pi\in\Phi_\Phi} \mathbb{E}_{\phi\sim P_\Phi} \left[ J^\pi(\phi)\right]\notin \Pi^\star_\Phi $. There then exists some set $\Phi'\subseteq\Phi$ with non-zero measure according to $P_\Phi$ such that:
	 \begin{align}
	 	J^{\pi^\dagger}(\phi)<J^{\pi^\star}(\phi),
	 \end{align}
	 for any $\phi\in\Phi'$ and $\pi^\star\in \Pi^\star_\Phi$. Taking expectations it follows:
	  \begin{align}
	 	\mathbb{E}_{\phi\sim P_\Phi}\left[J^{\pi^\dagger}(\phi)\right]<	\mathbb{E}_{\phi\sim P_\Phi}\left[J^{\pi^\star}(\phi)\right]\le \max_{\pi\in\Pi_\Phi}\mathbb{E}_{\phi\sim P_\Phi}\left[J^{\pi}(\phi)\right],
	 \end{align}
	 implying $\pi^\dagger(\cdot,\phi)\notin \argmax_{\pi\in\Phi_\Phi} \mathbb{E}_{\phi\sim P_\Phi} \left[ J^\pi(\phi)\right]$, which is a contradiction. As we have proved $\Pi^\star_\Phi=\argmax_{\pi\in\Pi_\Phi} \mathbb{E}_{\phi\sim P_\Phi} \left[ J^\pi(\phi)\right] $, we are left to show $\argmax_{\pi\in\Pi_\Phi} \mathbb{E}_{\phi\sim P_\Phi} \left[ J^\pi(\phi)\right]=\argmax_{\pi\in\Pi_\textrm{QBRL}} J^\pi_\textrm{Bayes}$.
	 \begin{align}
	 	\mathbb{E}_{\phi\sim P_\Phi} \left[ J^\pi(\phi)\right]=&	\mathbb{E}_{\phi\sim P_\Phi} \left[\mathbb{E}_{\tau_\infty\sim P_\infty^{\pi}(\phi)} \left[\sum_{t=0}^\infty r_t\right]\right],\\
	 	=&	\mathbb{E}_{\tau_\infty\sim P_\textrm{QBRL}^{\pi}} \left[\sum_{t=0}^\infty r_t\right],
	 \end{align}
	  where $ P_\textrm{QBRL}^{\pi}=\mathbb{E}_{\phi\sim P_\Phi} \left[P_\infty^{\pi}(\phi)\right]$. As $\pi(\phi,s_t)$ is context-conditioned, we can marginalise over all contexts using the posterior to obtain the predictive history-conditioned QBRL policy: $\pi_\textrm{QBRL}(h_t)=\mathbb{E}_{\phi\sim P_\Phi(h_t)}\left[ \pi(\phi,s_t)\right]$. It follows that $ P_\textrm{QBRL}^{\pi}= P^{\pi_\textrm{QBRL}}_\infty$ where $P^{\pi_\textrm{QBRL}}_\infty$ is the Bayesian predictive distribution over $h_\infty$ using context-conditioned policies with density:
	  		\begin{align}
	  			p^{\pi_\textrm{QBRL}}(h_\infty)=p_0(s_0)\prod_{i=0}^\infty \pi_\textrm{QBRL}(a_t\vert h_t)p(r_t,s_{t+1}\vert h_t),
	  		\end{align}
	  	hence:
	  \begin{align}
	  	\mathbb{E}_{\phi\sim P_\Phi} \left[ J^\pi(\phi)\right]=	\mathbb{E}_{\tau_\infty\sim P^{\pi_\textrm{QBRL}}_\infty} \left[\sum_{t=0}^\infty r_t\right]=J^{\pi_\textrm{QBRL}}_\textrm{Bayes},\\
	  	\implies \argmax_{\pi\in\Pi_\Phi} \mathbb{E}_{\phi\sim P_\Phi} \left[ J^\pi(\phi)\right]=\argmax_{\pi\in\Pi_\Phi} J^{\pi_\textrm{QBRL}}_\textrm{Bayes}.
	  \end{align}
	  Finally, as each $\pi(\cdot,\phi)\in\Pi_\phi$ indexes a $\pi_\textrm{QBRL}(\cdot,\phi)\in \Pi_\textrm{QBRL}$, 
	  \begin{align}
	  	\max_{\pi\in\Pi_\Phi} J^{\pi_\textrm{QBRL}}_\textrm{Bayes}=\max_{\pi\in\Pi_\textrm{QBRL}} J^{\pi}_\textrm{Bayes}\implies 	\Pi^\star_{\textnormal{\textrm{QBRL}}}=\argmax_{\pi\in\Pi_\textrm{QBRL}} J^\pi_\textrm{Bayes},
	  \end{align}
	where $\Pi^\star_{\textnormal{\textrm{QBRL}}}=\left\{\mathbb{E}_{\phi\sim P_\Phi(h_t)} \left[ \pi^\star(\cdot,\phi )\right]\vert \pi^\star(\cdot,\phi)\in \Pi^\star_\Phi\right\}$.
	\end{proof}
\end{theorem}

 \begin{corollary} \label{proof_app:counterexample_MDP} There exist MDPs with priors such that $\Pi_\textrm{QBRL}^\star \cap \Pi_\textrm{Bayes}^\star   =\varnothing$.
 	\begin{proof}
 		We consider the tiger problem as a counter example \citep{Kaelbling98} with $\gamma =0.9$, $r_\textrm{tiger}=-500$, $r_\textrm{gold}=10$ and $r_\textrm{listen}=-1$. Details of the space of MDPs can be found in \cref{app:tiger_problem}. We index the MDP with the tiger in the left door as $\phi=\textrm{tiger left}$ and the tiger in the right door as  $\phi=\textrm{tiger right}$. Consider the uniform prior over MDPs $P(\phi=\textrm{tiger left})=P(\phi=\textrm{tiger right})=0.5$. As agents always start in state $s_0$, it suffices to find the optimal MDP conditioned policies in $s_0$:
 		\begin{align}
 			\pi^\star(s_0, \phi=\textrm{tiger left}) =\delta(a=\textrm{open right}),\quad \pi^\star(s_0, \phi=\textrm{tiger right}) =\delta(a=\textrm{open left})
 		\end{align}
 		 From \cref{proof_app:bayes_value}, it follows that the optimal Bayesian contextual policy is a mixture of these two policies using the prior:
 		 \begin{align}
 		 	\pi^\star(s_0) =0.5(\delta(a=\textrm{open right})+\delta(a=\textrm{open left})).
 		 \end{align}
 	 	 The optimal $Q$-function for the optimal MDP-conditioned policies are
 	 	 	\begin{align}
 	 	 	&Q^\star(s_0,a=\textrm{open right}, \phi=\textrm{tiger left}) =Q^\star(s_0,a=\textrm{open left}, \phi=\textrm{tiger right}) = \frac{r_\textrm{gold}}{1-\gamma}=100,\\
 	 	 	&Q^\star(s_0,a=\textrm{open right}, \phi=\textrm{tiger right}) =Q^\star(s_0,a=\textrm{open left}, \phi=\textrm{tiger left}) \\
 	 	 	&\quad\quad\quad= r_\textrm{tiger}+\frac{\gamma r_\textrm{gold}}{1-\gamma}=-155.
 	 	 \end{align}
  	 Using \cref{proof_app:bayes_value}, we can find the Bayesian value function for the optimal contextual policy: 
  	 \begin{align}
  	 	&Q^\star_\textrm{QBRL}(s_0,a=\textrm{open right})\\
  	 	&\quad=0.5 \left( Q^\star(s_0,a=\textrm{open right}, \phi=\textrm{tiger left})+ Q^\star(s_0,a=\textrm{open right}, \phi=\textrm{tiger right}) \right),\\
  	 	&\quad=-27.5,\\
  	 	&Q^\star_\textrm{QBRL}(s_0,a=\textrm{open left})\\
  	 	&\quad=0.5 \left( Q^\star(s_0,a=\textrm{open left}, \phi=\textrm{tiger left})+ Q^\star(s_0,a=\textrm{open left}, \phi=\textrm{tiger right}) \right),\\
  	 	&\quad=-27.5,
  	 \end{align}
   from which the Bayesian return for the optimal contextual policy follows:
   \begin{align}
   	J^{\pi^\star_\textrm{QBRL}}_\textrm{Bayes}&= \mathbb{E}_{s \sim \delta(s_0)}\left[\mathbb{E}_{a \sim \pi^\star_\textrm{QBRL}(s)}\left[Q^\star_\textrm{QBRL}(s,a)\right]\right],\\
   	&= 0.5\left(Q^\star_\textrm{QBRL}(s_0,a=\textrm{open left})+Q^\star_\textrm{QBRL}(s_0,a=\textrm{open right})\right),\\
   	&=-27.5.
   \end{align}
Now consider the policy that always listens $\pi^\dagger=\delta(a=\textrm{listen})$. The Bayesian return for this policy is:
\begin{align}
	J^{\pi^\dagger}_\textrm{Bayes}=\frac{r_\textrm{listen}}{1-\gamma}=-10,
\end{align}
hence:
\begin{gather}
	J^{\pi^\star_\textrm{QBRL}}_\textrm{Bayes}< J^{\pi^\dagger}_\textrm{Bayes}\le\max_{\pi\in\Pi_\mathcal{H}} J^{\pi}_\textrm{Bayes}=J^{\pi^\star}_\textrm{Bayes},\\
	\implies \Pi_\textrm{QBRL}^\star \cap \Pi_\textrm{Bayes}^\star =\varnothing,
\end{gather}
as required. 
 	\end{proof}
\end{corollary}

\renewcommand{\thesection}{\Alph{section}}
\setcounter{section}{2}

\newpage

\section{Network Details}
 \begin{wrapfigure}{r}{0.31\textwidth}
	\centering
	\vspace{-0.4cm}
	\includegraphics[width=0.3\textwidth]{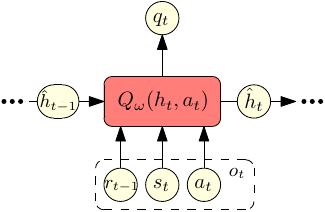}
	\caption{Recurrent $Q$ Network}
	\label{fig:recurrent_q_network}
	\vspace{-0.5cm}
\end{wrapfigure}Our recurrent $Q$ network architecture is shown \cref{fig:recurrent_q_network}. We encode the prior history-action pair via a recurrent variable $\hat{h}_{t}\coloneqq \{h_{t},a_{t}\}$. At each timestep our network inputs $\hat{h}_{t-1}$ and a tuple of observations $o_t\coloneqq\{r_{t-1},s_t,a_t\}$ and outputs the value variable $q_t=Q_\omega(h_t,a_t)=Q_\omega(\hat{h}_{t-1},o_t)$ and the new recurrent encoding $\hat{h}_t$. 
\subsection{Aleatoric Network}
\label{app:aleatoric}
 As discussed in \cref{sec:network_spec}, to model the distribution over optimal Bellman operators $P_B(h_t,a_t,\phi;\omega)$ we introduce a base variable $z_\textrm{al}\in\mathbb{R}$ with a tractable distribution $P_\textrm{al}$; in this paper we use a zero-mean, unit variance Gaussian $\mathcal{N}(0,1)$. We then generate $b_t$ using a change of variables $b_t=B(z_\textrm{al},q_t,\phi)$ parameterised by $\phi\in\Phi$,  where $B(z_\textrm{al},q_t,\phi)$ is a mapping in $z_\textrm{al}$ with inverse $
z_\textrm{al}={B}^{-1}(b_t,q_t,\phi)$. Under this change of variables, $\mathbb{E}_{b_t\sim P_B(h_t,a_t,\phi;\omega) }\left[ f(b_t)\right]=	\mathbb{E}_{z_\textrm{al} \sim P_\textrm{al}} \left[f\circ B(z_\textrm{al},q_t,\phi) \right]$
for any integrable $f:\mathbb{R}\rightarrow\mathbb{R}$. We refer to  $B(z_\textrm{al},q_t,\phi)$ as the \emph{aleatoric} network as it characterises the aleatoric uncertainty in the MDP and its expressiveness implicitly determines the space of MDPs that our model can represent. 

\begin{wrapfigure}{L}{0.31\textwidth}
		\vspace{-0.1cm}
			\centering
			\includegraphics[width=0.23\textwidth]{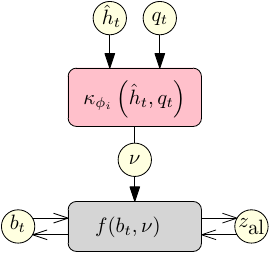}
			\caption{Details of Aleatoric Flow}
			\label{fig:aleatoric_flow}
		\vspace{-0.2cm}
\end{wrapfigure}
We define our aleatoric flow by adapting the autoregressive  flow \citep{Kingma16}. We take inputs from the RNN $Q$-function approximator outputs $\hat{h}_t,q_t$ (including the history encoding) and pass them through a conditioner $\kappa_{\phi_i}(\hat{h}_t,q_t)$, which is a feed forward neural network parametrised by $\phi_i$ where $\phi_i\subset \phi$. The output of the conditioner is a vector that defines the parameters for the coupling function. We use an inverse autoregressive flow, with $z_\textrm{al}\in \mathbb{R}^2$ followed with a dimensionality reduction layer to reduce the dimension of the output to 1 and abs layers, as detailed in \citet{Nielsen20}.  Since only $b_t=B(q_t,z_\textrm{al},\phi)$ needs to be bijective in $z_\textrm{al}$, there are also no restrictions on $Q_\omega(h_t,a_t)$, allowing us to use any arbitrary RNN.  The aleatoric network then consists of $L$ coupling functions in composition:
\begin{align}
	B(z_\textrm{al},q_t,\phi)=\ f_{L}\circ f_{L-1}\circ\cdots f_2\circ  f_1(z_\textrm{al},\hat{h}_t,q_t,\phi).
\end{align}

\subsection{Epistemic Network} 
\label{app:epistemic}
Given the aleatoric network $ B(z_\textrm{al},q_t,\phi)$, dataset of bootstrapped samples $\mathcal{D}_\omega(h_t)\coloneqq \{b_i\}_{i=0}^{t-1}$ and prior over parameters $P_\Phi$, our goal is to infer the posterior $P_\Phi(\mathcal{D}_\omega(h_t))$ to obtain the predictive mean:
\begin{align}
	\hat{B}[Q_\omega](h_t,a_t) \coloneqq \mathbb{E}_{\phi\sim P_\Phi(\mathcal{D}_\omega(h_t))}\left[ \mathbb{E}_{z_\textrm{al} \sim P_\textrm{al}}\left[B(q_t,z_\textrm{al},\phi)\right] \right].
\end{align}
Unfortunately, inferring the posterior and carrying out marginalisation exactly is intractable for all but the simplest aleatoric networks, which would not have sufficient capacity to represent a complex target  distribution $P_B(h_t,a_t,\phi;\omega)$. We instead look to variational inference using a normalising flow to learn a tractable approximation.

Like in \cref{app:aleatoric}, we introduce a base variable $z_\textrm{ep}\in\mathbb{R}^d$ with a tractable distribution $P_\textrm{ep}$; again we use a zero-mean Gaussian $\mathcal{N}(0,I^d)$. We then make a transformation of variables $\phi = t_\psi(z_\textrm{ep})$ where  $t_\psi:\mathbb{R}^d\rightarrow \mathbb{R}^d$ is a bijective mapping parametrised by $\psi\in\Psi$ with inverse $z_\textrm{ep}= t_\psi^{-1}(\phi)$. We refer to	 $t_\psi(z_\textrm{ep})$ as the epistemic network as it characterises the epistemic uncertainty in $\phi$. From the change of variables formula, it follows that the resulting variational distribution $P_\psi$  has a density $p_\psi(\phi)=\left\lvert\det\left(J_\psi(\phi)\right)\right\rvert p_\textrm{ep} \circ t_\psi^{-1}(\phi)$
where $J_\psi(\phi)\coloneqq \nabla_\phi t_\psi^{-1}(\phi)$ is the Jacobian of the inverse mapping. Using variational inference, we treat $P_\psi$ as an approximation of the true posterior $P_\Phi(h_t)$, which we learn by minimising the KL-divergence between the two distributions $\kl{P_\psi}{P_\Phi(h_t)}$. This is equivalent to minimising the following negative evidence lower-bound (ELBO) objective with respect to $\psi$:

\begin{align}
	\textrm{ELBO}(\psi; h_t,\omega)\coloneqq\mathbb{E}_{z_\textrm{ep}\sim P_{\textrm{ep}}}\left[\left( \sum_{i=0}^{t-1}\left(B^{-1}(b_i,q_i,\phi)^2-\log \left\lvert\partial_{b}B^{-1}(b_i,q_i,\phi)\right\rvert \right)- \log p_\Phi(\phi) \right)\bigg\vert_{\phi= t_{\psi}(z_\textrm{ep})}\right].
\end{align}

\begin{figure}
	\centering
	\includegraphics[scale=0.85]{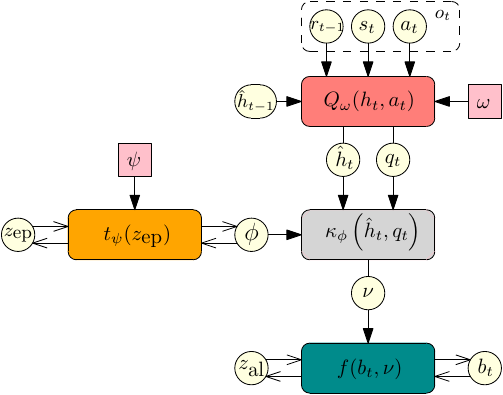}
	\caption{Schematic of \textsc{ben}}
	\label{fig:ben_network}
	\vspace{-0.2cm}
\end{figure}

We provide a fully connected schematic of \textsc{ben} with flow details in  \cref{fig:ben_network}.

To derive this result, we start with the definition of the KL-divergence $\kl{P_\psi}{P_\Phi(h_t)}$, using Bayes' rule to re-write the log-posterior:
\begin{align}
	\kl{P_\psi}{P_\Phi(h_t)}\coloneqq& \mathbb{E}_{\phi\sim P_\psi(\phi)} \left[ \log p_\psi(\phi) - \log p_\Phi(\phi\vert h _t) \right],\\
	=& \mathbb{E}_{\phi\sim P_\psi(\phi)} \left[ \log p_\psi(\phi) - \log p_\Phi(h_t\vert \phi) -  \log p_\Phi( \phi)+ \log p(h_t)  \right],\\
	=& \mathbb{E}_{\phi\sim P_\psi(\phi)} \left[ \log p_\psi(\phi) - \log p_\Phi(h_t\vert \phi) -  \log p_\Phi( \phi)  \right] .
\end{align}
As the log-evidence $\log p(h_t)$ has no dependence on $\psi$, we can omit it from our objective, instead maximising the ELBO: 
\begin{align}
	\textrm{ELBO}(\psi; h_t,\omega)=  \mathbb{E}_{\phi\sim P_\psi(\phi)} \left[ \log p_\Phi(h_t\vert \phi) + \log p_\Phi( \phi)  - \log p_\psi(\phi) \right].
\end{align}
We write the expectation with respect to $P_\psi$ using the transformation of variables:
\begin{align}
	\textrm{ELBO}(\psi; h_t,\omega)&=  \mathbb{E}_{z_\textrm{ep}\sim P_\textrm{ep}(z_\textrm{ep})} \left[ \log p_\Phi(h_t\vert t_\psi(z_\textrm{ep}) + \log p_\Phi( t_\psi(z_\textrm{ep})  - \log p_\psi(t_\psi(z_\textrm{ep}) \right],\\
	&=  \mathbb{E}_{z_\textrm{ep}\sim P_\textrm{ep}(z_\textrm{ep})} \left[ \log p_\Phi(h_t\vert t_\psi(z_\textrm{ep}) \right]+\mathbb{E}_{z_\textrm{ep}\sim P_\textrm{ep}(z_\textrm{ep})} \left[  \log p_\Phi( t_\psi(z_\textrm{ep})  - \log p_\psi(t_\psi(z_\textrm{ep}) \right].
\end{align}
Now, by the definition of $	p_\psi(\phi)$, it follows:
\begin{align}
	p_\psi\circ t_\psi(z_\textrm{ep})&=\left\lvert\det\left(J_\psi\circ  t_\psi(z_\textrm{ep})\right)\right\rvert p_\textrm{ep} \circ t_\psi^{-1}\circ  t_\psi(z_\textrm{ep}),\\
	&=\left\lvert\det\left(\nabla_\phi t_\psi^{-1}\circ  t_\psi(z_\textrm{ep})\right)\right\rvert p_\textrm{ep} (z_\textrm{ep}),\\
	& =p_\textrm{ep} (z_\textrm{ep}),
\end{align}
hence as $p_\textrm{ep} (z_\textrm{ep})$ has no dependence on $\psi$, we can omit it from the objective, yielding:
\begin{align}
	\textrm{ELBO}(\psi; h_t,\omega)&=  \mathbb{E}_{z_\textrm{ep}\sim P_\textrm{ep}(z_\textrm{ep})} \left[ \left( \log p_\Phi(h_t\vert \phi) + \log p_\Phi( \phi)  \right)\vert_{\phi= t_\psi(z_\textrm{ep})}\right],\\
	&= \mathbb{E}_{z_\textrm{ep}\sim P_\textrm{ep}(z_\textrm{ep})} \left[\left(\log\left(\prod _{i=0}^{t-1} p_B(b_i\vert h_i,a_i,\phi;\omega)\right)+\log p_\Phi(\phi)\right)\bigg\vert_{\phi=t_\psi(z_\textrm{ep})} \right],\\
	&= \mathbb{E}_{z_\textrm{ep}\sim P_\textrm{ep}(z_\textrm{ep})} \left[\left(\sum_{i=0}^{t-1}\log p_B(b_i\vert h_i,a_i,\phi;\omega)+\log p_\Phi(\phi)\right)\bigg\vert_{\phi=t_\psi(z_\textrm{ep})} \right].
\end{align}
Finally we can derive the exact form of the $\log$-density $p_B(b_i\vert h_i,a_i,\phi;\omega)$ using the change of variables formula under $ b_t=B(z_\textrm{al},q_t,\phi)$:
\begin{align}
	\log p_B(b_i\vert h_i,a_i,\phi;\omega)=\log \left(\exp(-B^{-1}(b_i,q_i,\phi)^2) \left\lvert\partial_{b}B^{-1}(b_i,q_i,\phi)\right\rvert \right),\\
	=-B^{-1}(b_i,q_i,\phi)^2+\log \left\lvert\partial_{b}B^{-1}(b_i,q_i,\phi)\right\rvert.
\end{align}
Substituting and multiplying by $-1$, thus changing to an objective to minimise rather than maximise, yields our desired result:
\begin{align}
	\textrm{ELBO}(\psi; h_t,\omega)\coloneqq\mathbb{E}_{z_\textrm{ep}\sim P_{\textrm{ep}}}\left[\left( \sum_{i=0}^{t-1}\left(B^{-1}(b_i,q_i,\phi)^2-\log \left\lvert\partial_{b}B^{-1}(b_i,q_i,\phi)\right\rvert \right)- \log p_\Phi(\phi) \right)\bigg\vert_{\phi= t_{\psi}(z_\textrm{ep})}\right].
\end{align}

\subsection{Network Training}
\label{app:nework_training}
 \paragraph{Prior Initialisation}
\begin{wrapfigure}{R}{0.5\textwidth} \vspace{-0.5cm}
	\begin{minipage}{0.5\textwidth}
		\vspace{-0.2cm}
		\begin{algorithm}[H]
			\caption{$\textsc{PriorInitialisation}(P_\Phi,s_0,\mathcal{D}_\textrm{prior})$}
			\label{alg:approxBRL_prior}
			\begin{algorithmic}
				\STATE Initialise $\omega$
				\FOR{ $N_\textrm{Pretrain}$ steps}
				\STATE Sample action $a\sim \rho$
				\STATE Sample two MDPs $\phi,\phi'\sim P_\Phi$
				\STATE Sample aleatoric variables $z_\textrm{al},z_\textrm{al}'\sim P_\textrm{al}$
				\STATE $q_0 = Q_\omega(s_0,a)$
				\STATE $\omega \leftarrow \omega -\alpha (B(q_0, z_\textrm{al},\phi)-q_0)\nabla_{\omega}(B(q_0, z_\textrm{al}',\phi')-q_0)$
				\STATE Sample from prior data $s,a\sim \mathcal{D}_\textrm{prior}$				\STATE Sample two MDPs $\phi,\phi'\sim P_\Phi$ 
				\STATE Sample rewards-state transition $r,s_+\sim P_{R,S}(s,a,\phi)$
				\STATE Sample rewards-state transition  $r',s'_+\sim P_{R,S}(s,a,\phi')$
				\STATE $\omega \leftarrow \omega -\alpha (r+\gamma \max_{a'\in\mathcal{A}} Q_\omega(s,a,r,s_+,a')-Q_\omega(s,a))\nabla_{\omega}(\gamma \max_{a'\in\mathcal{A}} Q_\omega(s,a,r',s_+',a')-Q_\omega(s,a))$
				\ENDFOR
			\end{algorithmic}
		\end{algorithm}
		\vspace{-0.6cm}
	\end{minipage}
\end{wrapfigure}
Before any actions have been taken, we can minimise the MSBBE using the initial state $s_0$ and the prior $P_\phi$, which we assume is tractable to obtain samples from. This has the advantage of initialising the $Q$-function approximator to incorporate any prior domain knowledge we have about the MDP, in addition to ensuring that an optimal Bayesian Bellman equation is approximately satisfied before training starts. Once the posterior is updated using a new observation, we shouldn't expect the solution to the MSBBE to change significantly to reflect the updated belief. Finally, there may be prior knowledge about state and reward transitions that are available to us a priori that we would like to encode in the $Q$-function approximator. If an agent in state $s$ taking action $a$ always transitions according to a known conditional distribution $P_{R,S}(s,a,\phi)$, then we can use this information to solve the Bayesian Bellman equation conditioned on $s,a$. We combine all such state-action pairs into a dataset $\mathcal{D}_\textrm{prior}\coloneqq \{s_i,a_i\}_{i=1}^{K_\textrm{prior}}$, for which we minimise the MSSBE:
\begin{align}
	\mathcal{L}(\omega;\mathcal{D}_\textrm{prior})=\sum_{i=1}^{K_\textrm{prior}} \left(\mathbb{E}_{\phi\sim P_\Phi}\left[\mathbb{E}_{r_i,s_{i+1}\sim P_{R,S}(s_i,a_i,\phi)}\left[r+\gamma \max_{a'\in\mathcal{A}}Q_\omega(s_i,a_i,r_i,s_{i+1},a')\right]\right]-Q_\omega(s_i,a_i)\right)^2.
\end{align}
We give specific details of $\mathcal{D}_\textrm{prior}$ in the context of our search and rescue environment in \cref{app:search_rescue_problem}. Both MSBBE objectives can be minimised using stochastic gradient descent with two independent samples from the prior to avoid bias in our updates, as outlined in \cref{alg:approxBRL_prior}. Note that for domains where we don't have such knowledge, we can take $\mathcal{D}_\textrm{prior}=\varnothing$ and ignore the minimisation steps on $\mathcal{L}(\omega;\mathcal{D}_\textrm{prior})$. \textsc{ben}'s incorporation of prior knowledge does not require a full generative model of the environment dynamics and demonstrations can be from simulated or related MDPs that do not exactly match the set of tasks the agent is in at test time.

 \paragraph{Posterior Updating}
\begin{figure}[H]

	\centering
	\includegraphics[scale=0.85]{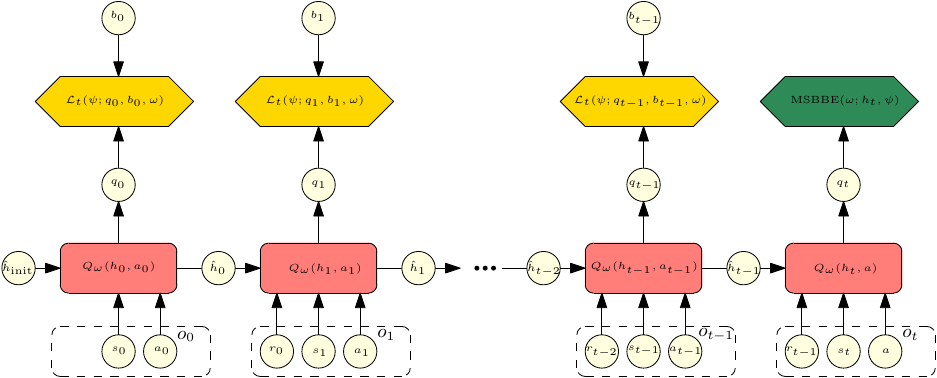}
	\caption{Schematic of \textsc{ben} Training Regime. Losses are shown as hexagons.}
	\label{fig:training_fig}

\end{figure}
To obtain an efficient algorithm, we note that the ELBO objective can be written as a summation: $\textrm{ELBO}(\psi; h_t,\omega)= \sum_{i=0}^{t-1}\mathcal{L}_t(\psi; q_i,b_i,\omega)$,
where each sub-objective is:
\begin{align}
	\mathcal{L}_t(\psi; q_i,b_i,\omega)\coloneqq\mathbb{E}_{z_\textrm{ep}\sim P_{\textrm{ep}}}\left[\left(B^{-1}(b_i,q_i,\phi\right)^2-\log \left\lvert\partial_{b}B^{-1}(b_i,q_i,\phi)\right\rvert -\frac{1}{t} \log p_\Phi(\phi) \bigg\vert_{\phi= t_{\psi}(z_\textrm{ep})}\right].
\end{align} 
\begin{wrapfigure}{L}{0.5\textwidth}
	\begin{minipage}{0.5\textwidth}
		\vspace{-1cm}
		\begin{algorithm}[H]
			\caption{$\textsc{PosteriorUpdating}(h_t,\psi,\omega)$}
			\label{alg:approxBRL_posterior}
			\begin{algorithmic}
				\FOR{$N_\textrm{Update}$ steps} 
				\STATE $\hat{h}\leftarrow \hat{h}_\textrm{init}$
				\STATE $o\leftarrow \{s_0,a_0\}$
				\FOR{$i\in[0:t-1]$}
				\STATE $\hat{h},q_i \leftarrow Q_\omega(\hat{h},o)$
				\STATE $b_i \leftarrow r_i +\gamma \max_{a'} Q_\omega(\hat{h},o,r_i, s_{i+1},a')$
				\STATE $g_\psi\sim \nabla_\psi\mathcal{L}_t(\psi;q_i,b_i,\omega)$
				\STATE $\psi\leftarrow \psi -\alpha_\psi g_\psi$
				\STATE $o \leftarrow r_i,s_{i+1},a_{i+1}$
				\ENDFOR
				\STATE $q_t \leftarrow Q_\omega(\hat{h},o)$ 
				\STATE $\beta_\omega\sim \nabla_\omega\textrm{MSBBE}(\omega; q_t,\psi)$
				\STATE $\omega\leftarrow  \omega -\alpha_\omega \beta_\omega$
				\ENDFOR
			\end{algorithmic}
		\end{algorithm}
		\vspace{-0.6cm}
	\end{minipage}
\end{wrapfigure}As shown in \cref{fig:training_fig}, we can minimise $\textrm{ELBO}(\psi; h_t,\omega)$ by unrolling the RNN, starting at $i=0$. After each timestep, we obtain $q_i$, which can be used to minimise the loss $	\mathcal{L}_t(\psi; q_i,b_i,\omega)$ with the observation $b_i$ whilst keeping $\omega$ fixed. Once the network has been unrolled to the timestep $t$, we can use the output to minimise the MSBBE. Like in for our prior initialisation algorithm in \cref{alg:approxBRL_prior}, it is important that we sample two independent samples $\phi,\phi' \sim P_\phi(h_t)$ from our approximate posterior when minimising the MSBBE to avoid biased gradient estimates.  Once $t$ becomes too large, we can truncate the sequences to length $t'$, starting at state $s_{t-t'}$ instead of $s_0$. Like when target parameters used to stabilise frequentist TD methods \citep{Fellows23}, this updating ensures that the $Q$-network is updated on an asymptotically slower timescale to the posterior parameters, and we tune the length of truncation $t'$ for the sequence and stepsizes $\alpha_\psi$ and $\alpha_\omega$ to ensure stability.

\section{Experiments}
\label{app:experiments}

\subsection{Tiger Problem}
\label{app:tiger_problem}

\begin{figure}[H]
	\centering
	\includegraphics[scale=1.2]{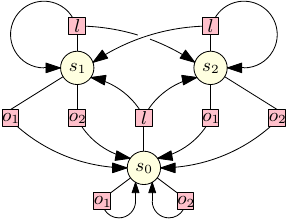}
	\caption{Tiger Problem MDP}
	\label{fig:tiger_problem}
\end{figure}
The aim of this empirical evaluation is to verify our claim that \textsc{ben} can learn a Bayes-optimal policy and compare \textsc{ben} to existing model-free approaches. We evaluate \textsc{ben} in the counterexample tiger problem domain from \cref{proof:counterexample_MDP}, which allows for comparison against a true Bayes-optimal policy. We show our tiger problem MDP in \cref{fig:tiger_problem}. The agent is always initialised in state $s_0$ and can chose to open door 1 ($o_1$), open door 2 ($o_2$) or to listen ($l$): $\mathcal{A}\coloneqq \{o_1, o_2,l\}$. There are two possible MDPs the agent can be in, with the tiger assigned to either door 1 or door 2 randomly and the gold to the other door. If the agent chooses $o_1$, door 1 is opened and the agent receives a reward of $r_\textrm{tiger} = -500$ if the tiger is behind the door or $r_\textrm{gold} = 10$ if the gold is behind the door. The agent always transitions to state $s_0$ after selecting $o_1$ or $o_2$. If the agent chooses to listen, it receives a small negative reward of $r_\textrm{listen}=-1$ and  if the tiger is behind door $1$ transitions to state $s_1$ with probability $0.85$ and state $s_2$ with probability $0.1$, or if the tiger is behind door $2$, the agent transitions to state $s_2$ with probability $0.85$ and state $s_1$ with probability $0.1$.

\subsection{Tiger Problem Implementation Details}
We initialise the agent with a uniform prior over the two MDPs. The posterior for this problem is tractable so we use that in place of the epistemic network:
\begin{align}
	P_\Phi(\phi=\textrm{tiger in 1}\vert h_t) &= \frac{0.85^{N_1}\cdot0.1^{N_2}}{0.85^{N_1}\cdot0.1^{N_2}+0.1^{N_1}\cdot0.85^{N_2}},\\
	P_\Phi(\phi=\textrm{tiger in 2}\vert h_t) &= 1- P_\Phi(\phi=\textrm{tiger in 1}) 
\end{align}
where $N_1$ is the number of visitations to state $s_1$ and  $N_2$ is the number of visitations to state $s_2$. If the agent opens the door, the posterior trivially becomes:
\begin{align}
	P_\Phi(\phi\vert h_t) = \begin{cases}
\delta(\phi=\textrm{tiger in 1}),\quad a_{t-1}=o_1,r_{t-1}= r_\textrm{tiger},\\
\delta(\phi=\textrm{tiger in 1}),\quad a_{t-1}=o_2,r_{t-1}= r_\textrm{gold},\\
\delta(\phi=\textrm{tiger in 2}),\quad a_{t-1}=o_2,r_{t-1}= r_\textrm{tiger},\\
\delta(\phi=\textrm{tiger in 2}),\quad a_{t-1}=o_1,r_{t-1}= r_\textrm{gold}.
	\end{cases}
\end{align}
The aleatoric network can be handcoded as the pushforward of known transition distributions. We  vary the number of steps for the MSBBE minimisation with a learning rate of 0.02 using ADAM for the stochastic gradient descent. For the for $Q$-function approximator, we use a fully connected linear layer with ReLU activations, a gated recurrent unit and a final fully connected linear layer with ReLU activations. All hidden dimensions are 32. The dimension of $\hat{h}_0$ is 2. The input dimension is 1 + 2 =3 and the network output is 3 dimensional to reflect the three possible actions the agent can take. 

\begin{wrapfigure}{r}{0.4\textwidth}
	\vspace{-1.2cm}
	\centering
	\includegraphics[width=0.4\textwidth]{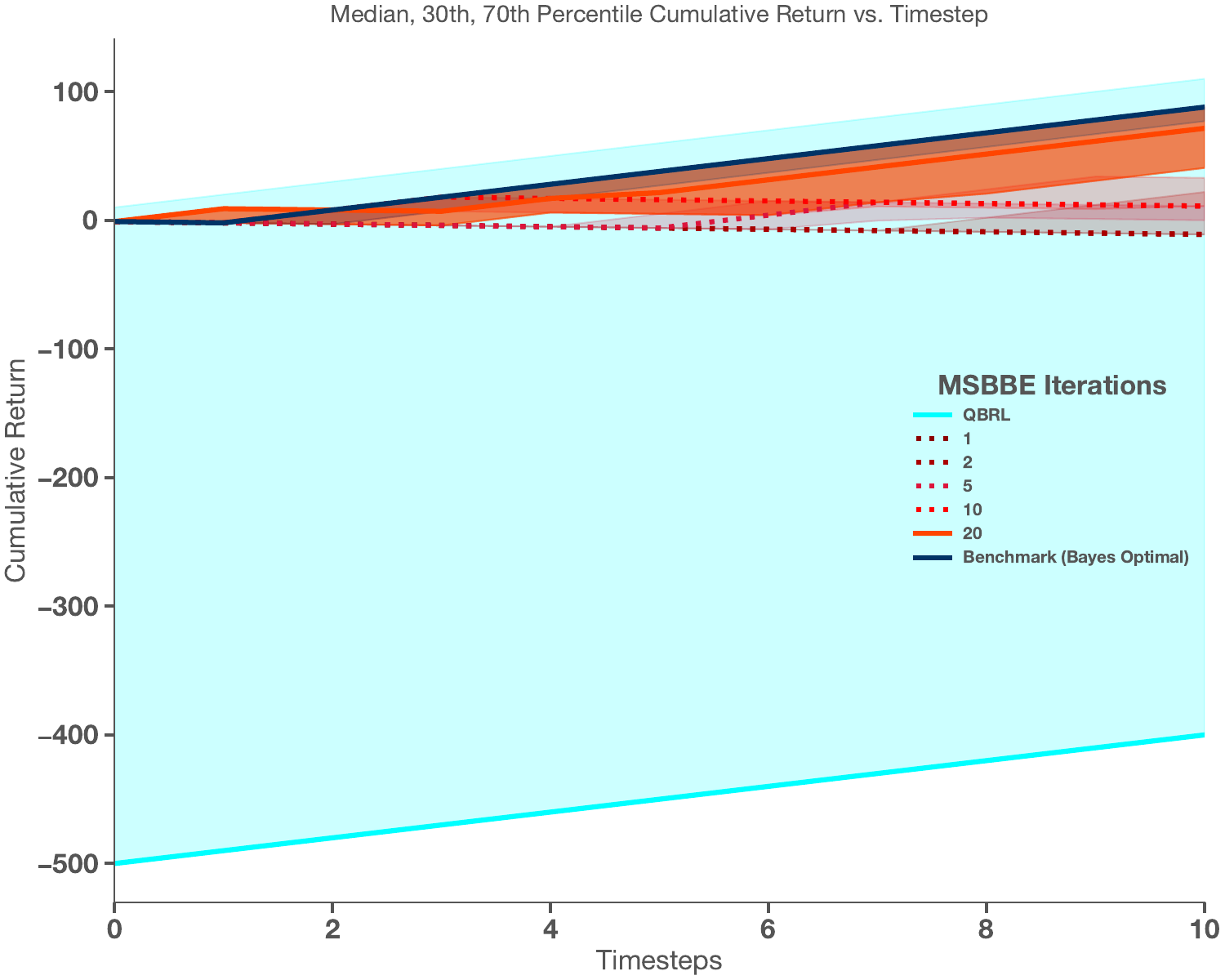}
	\caption{Results of evaluation in tiger problem showing \textsc{ben} with increasing minimisation steps on MSBBE vs Bayes-optimal and contextual oracles}
	\label{fig:tiger_results}
	\vspace{-0.3cm}
\end{wrapfigure}

\subsection{Tiger Problem Results}
 We initialise all agents in a tabula rasa setting with a uniform prior over MDPs and plot the median returns after each timestep in \cref{fig:tiger_results} for 11 timesteps, averaged over MDPs, each drawn uniformly.  We plot the performance of \textsc{ben} for a varying number of SGD minimisation steps on our MSBBE objective. \cref{fig:tiger_results} shows that by increasing the number of SGD minimisation steps, \textsc{ben}'s performance approaches that of the Bayes-optimal oracle and the variance in the policies decreases, with near Bayes-optimal performance attainted using 20 minimisation steps. We also compare \textsc{ben} to an oracle that is optimal over the space contextual policies, $\Pi^\star_\textrm{QBRL}$, which is an optimal policy for existing model-free approaches. As expected, the contextual optimal policy is limited to a mixture of optimal policies conditioned on $\phi$, hence the performance is comparatively poor: median returns are significantly lower than \textsc{ben} as contextual policies sample an initial action uniformly before acting optimally once the true MDP is revealed.

\subsection{Search and Rescue Problem}
\label{app:search_rescue_problem}
\begin{figure}[H]
	\centering
	\includegraphics[scale=1]{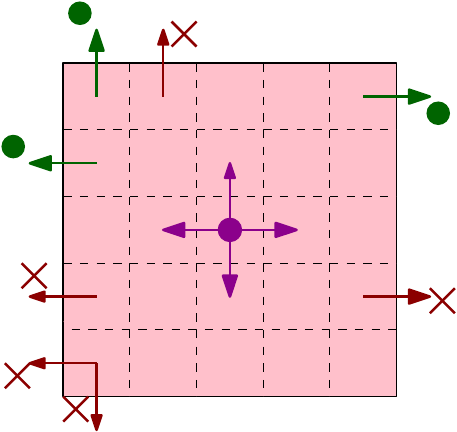}
	\caption{$5\times5$ Search and Rescue Problem MDP with 5 Hazards (red crossed) and 3 Victims (green circles). Agent (purple circle) is shown in $s_0$. Green actions yield reward $r_\textrm{rescue}$ and red actions yield reward $r_\textrm{hazard}$.}
	\label{fig:search_rescue_problem}
\end{figure}
We now present a novel search and rescue MDP designed to present a challenging extension to the toy tiger problem domain. An agent is tasked with rescuing $N_\textrm{victims}$ victims from a dangerous situation whilst avoiding any one of $N_\textrm{hazards}$ hazards. The agent's action space is $\mathcal{A}=\{\textrm{up}, \textrm{down}, \textrm{left},\textrm{right}, \textrm{listen}\}$. The agent can move in an $N_\textrm{grid}\times N_\textrm{grid}$ gridworld where $ N_\textrm{grid}$ is an odd number and transitions one square deterministically in the direction of the action taken. If the agent selects an action that would take it off the grid, it remains put and opens the door adjacent to its square in the direction of the action. If the agent opens a door with a victim behind, it receives a reward of $r_\textrm{victim}=10$ and the victim is removed from the MDP. If the agent . If the agent opens a door with a hazard behind, it receives a reward of $r_\textrm{hazard}=-100$ and the hazard remains in the MDP. We show an example MDP in \cref{fig:search_rescue_problem}.

 The agent is initialised in position $(0,0)$, which is the central square of the grid. The agent observes state $s\in\mathcal{S}\subset\mathbb{R}^{2+N_\textrm{victims} + N_\textrm{hazards}}$ where $l_\textrm{agent}\coloneqq (s_0,s_1)$ is the agent's location relative to $(0,0)$. The agent does not directly observe which doors have hazards or victims behind. If the agent chooses the action listen, their location remains put and they transition to a new state $s'$ where $s_i'$ for each $i\in \{2:1+N_\textrm{victims} + N_\textrm{hazards}\}$ is given by: 
 \begin{align}
 s_i' = \exp\left(-\frac{\lVert l_\textrm{agent}-l_i\rVert^2}{N_\textrm{grid}}+\eta \right),\quad \eta\sim \mathcal{N}(0;\sigma_\textrm{noise}^2)
 \end{align}
which a noisy variable correlated to the distance between the agent and each victim/hazard $l_i$. The victim locations are $\{l_i\}_{i\in\{2:N_\textrm{victims}+1\}}$ and the hazard locations are $\{l_i\}_{i\in\{N_\textrm{victims}+2:1+ N_\textrm{victims}+N_\textrm{hazard}\}}$. For each MDP, the victims and hazards are randomly assigned a square each adjacent to the grid and the initialised uniformly across that square. If an agent opens a door with a victim, their location becomes $(N_\textrm{grid}\cdot1000,N_\textrm{grid}\cdot1000)$ and no further reward can be obtained for that victim. Agents receive a small negative reward for listening $r_\textrm{listen}=-1$ and no reward for traversing the grid.

\subsection{Exploiting Prior Knowledge}
\label{app:search_and_rescue_prior}
For the search and rescue environment, there is domain knowledge that we can use to form $\mathcal{D}_\textrm{prior} $ that is common to all MDPs. The first example of this knowledge is that movement transitions are deterministic and yield no reward when the agent is traversing the grid. To make this precise, we define the set of states in the interior of the grid:
\begin{align}
	\mathcal{S}_\textrm{interior}\coloneqq \left\{s\vert \lvert s_0 \rvert < \frac{N_\textrm{grid} -1}{2},\lvert s_1 \rvert < \frac{N_\textrm{grid} -1}{2}, \{s_i \}_{i\ge2}=0\right\},
\end{align}
that is their location is not adjacent to the grid's boundary. All other values $s_2 : s_{1+N_\textrm{victims}+N_\textrm{hazards}}$ are set to 0. We define the set of movement actions to be $\mathcal{A}_\textrm{movement}\coloneqq \{up, down, left, right\}$. Taking an action $a\in \mathcal{A}_\textrm{movement}$ when in state $s\in \mathcal{S}_\textrm{interior}$ always moves the agent in the direction of the action selected without changing the other states and receives a reward of $0$, that is:
\begin{align}
P_S(s\in \mathcal{S}_\textrm{interior},a= up) &= \delta(s_1' = s_1 +1, s'_{\ne 1}=s_{\ne 1}  ) ,\\
P_S(s\in \mathcal{S}_\textrm{interior},a= down) &= \delta(s_1' = s_1 -1, s'_{\ne 1}=s_{\ne 1}  ) ,\\
P_S(s\in \mathcal{S}_\textrm{interior},a= right) &= \delta(s_0' = s_0 +1, s'_{\ne 0}=s_{\ne 0}  ) ,\\
P_S(s\in \mathcal{S}_\textrm{interior},a= left) &= \delta(s_0' = s_0 -1, s'_{\ne 0}=s_{\ne 0}  ),\\
P_R(s\in \mathcal{S}_\textrm{interior},a\in\mathcal{A}_\textrm{movement}) &= \delta(r=0 ),
\end{align}
allowing us to sample from $\mathcal{A}_\textrm{movement}\times \mathcal{S}_\textrm{interior}\subset\mathcal{D}_\textrm{prior}$ and apply the above transformation.

In addition to the deterministic transitions, we can also include prior reward information. Firstly we define the boundary states where the agent is adjacent to the edge of the grid
\begin{align}
		\mathcal{S}_\textrm{boundary}\coloneqq \left\{s\vert \lvert s_0 \rvert = \frac{N_\textrm{grid} -1}{2}, \{s_i \}_{i\ge2}=0\right\} \cup \left\{s\vert \lvert\lvert s_1 \rvert =\frac{N_\textrm{grid} -1}{2}, \{s_i \}_{i\ge2}=0\right\}
\end{align}
We note that again all non-locations states $s_2 : s_{1+N_\textrm{victims}+N_\textrm{hazards}}$ are set to $0$ because other values are specific to each MDP.  As agents and hazards are initialised uniformly in the squares adjacent to the grid, if an agent is in $\mathcal{S}_\textrm{boundary}$ and takes an action to move out of the grid (i.e. open a door), then the expected reward will be:
\begin{align}
	r_\textrm{prior}\coloneqq \frac{N_\textrm{victims}}{4 \times N_\textrm{grid}} \cdot r_\textrm{victim}+\frac{N_\textrm{hazards}}{4 \times N_\textrm{grid}} \cdot r_\textrm{hazard}.
\end{align}
\paragraph{Listening Information}
Key to solving the search and rescue environment is learning to listen before acting.

\subsection{Search and Rescue Implementation Details}
The epistemic network consists of two layers of ActNorm, a Masked Autoregressive Flow with two blocks and a LU linear decomposition and permutation as in \citet{dinh2016density}. The base distribution is a unit Gaussian. This takes the number of parameters in the Aleatoric Network is a projection to 2d space from 1d, an Inverse Autoregressive Flow, a LU Linear decomposition and Permutation, a projection back to 1d with the Slice Flow, and an Abs Flow.  The base distribution is a standard 1d Gaussian. The AbsFlow consists of 6 applications of the conditioner network, (K=6), and two layers.  We vary the number of steps for the MSBBE minimisation with a learning rate of 1e-4 using ADAM for the stochastic gradient descent and use a separate ADAM optimiser, with a learning rate of 1e-4 for the Epistemic Network training on the ELBO. For the for $Q$-function approximation, we use a fully connected linear layer with ReLU activations, a gated recurrent unit and a final fully connected linear layer with ReLU activations. All hidden dimensions are 64. The dimension of the hidden state $\hat{h}_0$ is 64.  The input size is the state space size 14 (4 number of victims + 8 number of hazards + 1 + 1 for x and y dims)  The input dimension is state space size + 1 for reward + 1 for action = 16. The network output is 5 dimensional to reflect the five possible actions the agent can take. The input to the conditioner network is number of aleatoric parameters+  hidden dim size + 1 for q value, and the hidden layers and output size are the number of aleatoric parameters. Only a subset of these aleatoric parameters are used as needed in each layer and the rest are dropped.

\subsection{Ablations}
\label{sec:ablations}
We carry out the following ablations in the zero-shot setting for the search and rescue environment, averaged over 7 seeds in this zero-shot test and plot the sample standard errors:

\paragraph{QBRL Approaches}

\begin{figure}[t!]
		\vspace{-1cm}
	\subfloat[Victims Saved Contextual Ablation]{%
		\includegraphics[width=.47\linewidth]{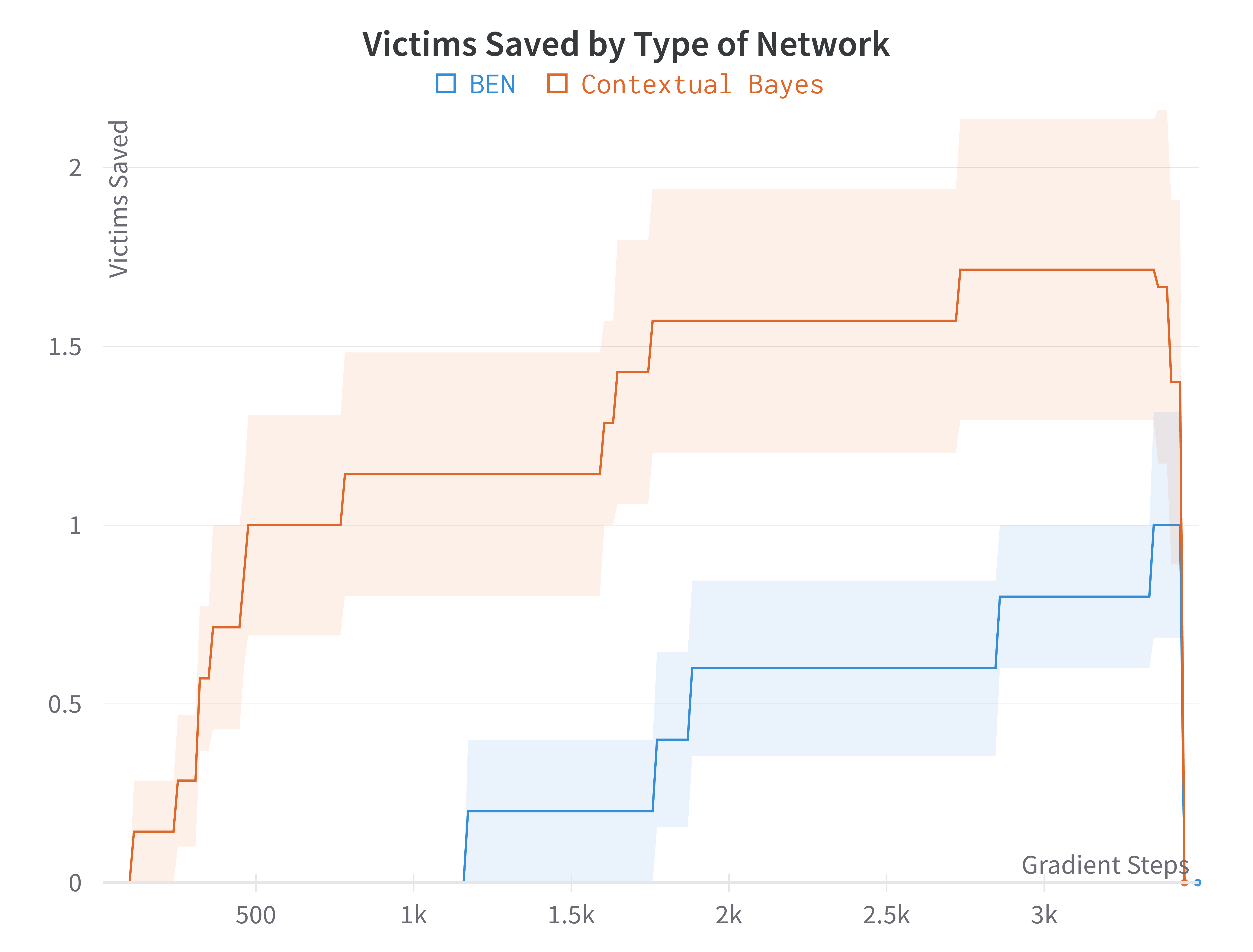}%
		\label{subfig:victims}%
	}\hfill
	\subfloat[Hazards Hit Contextual Ablation]{%
		\includegraphics[width=.47\linewidth]{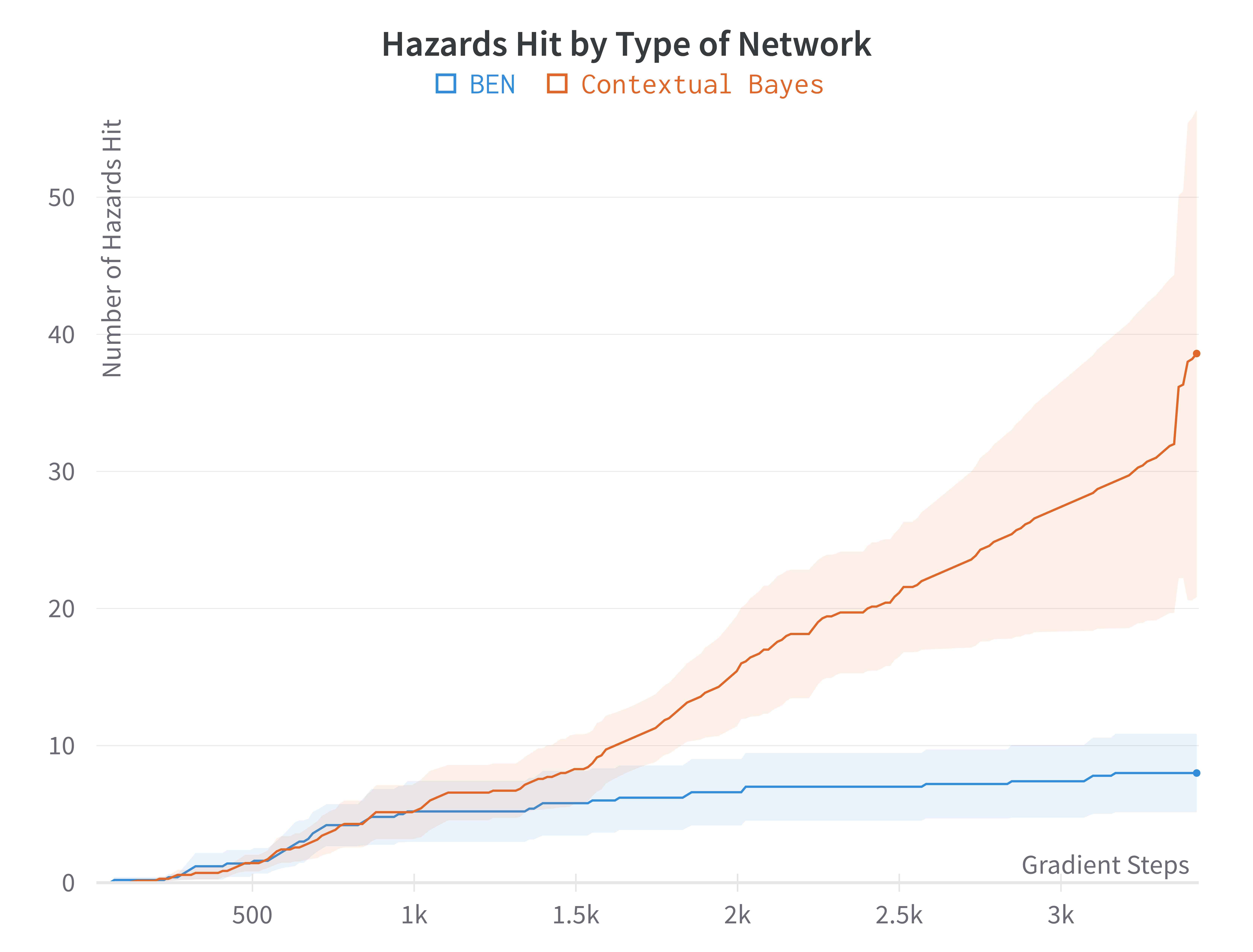}%
		\label{subfig:hazards}%
	}
	\label{fig:contextual}
	\caption{Contextual vs \textsc{ben} Ablation}
\end{figure}

\begin{figure}[h!]
	\subfloat[Return Aleatoric Ablation]{%
		\includegraphics[width=.47\linewidth]{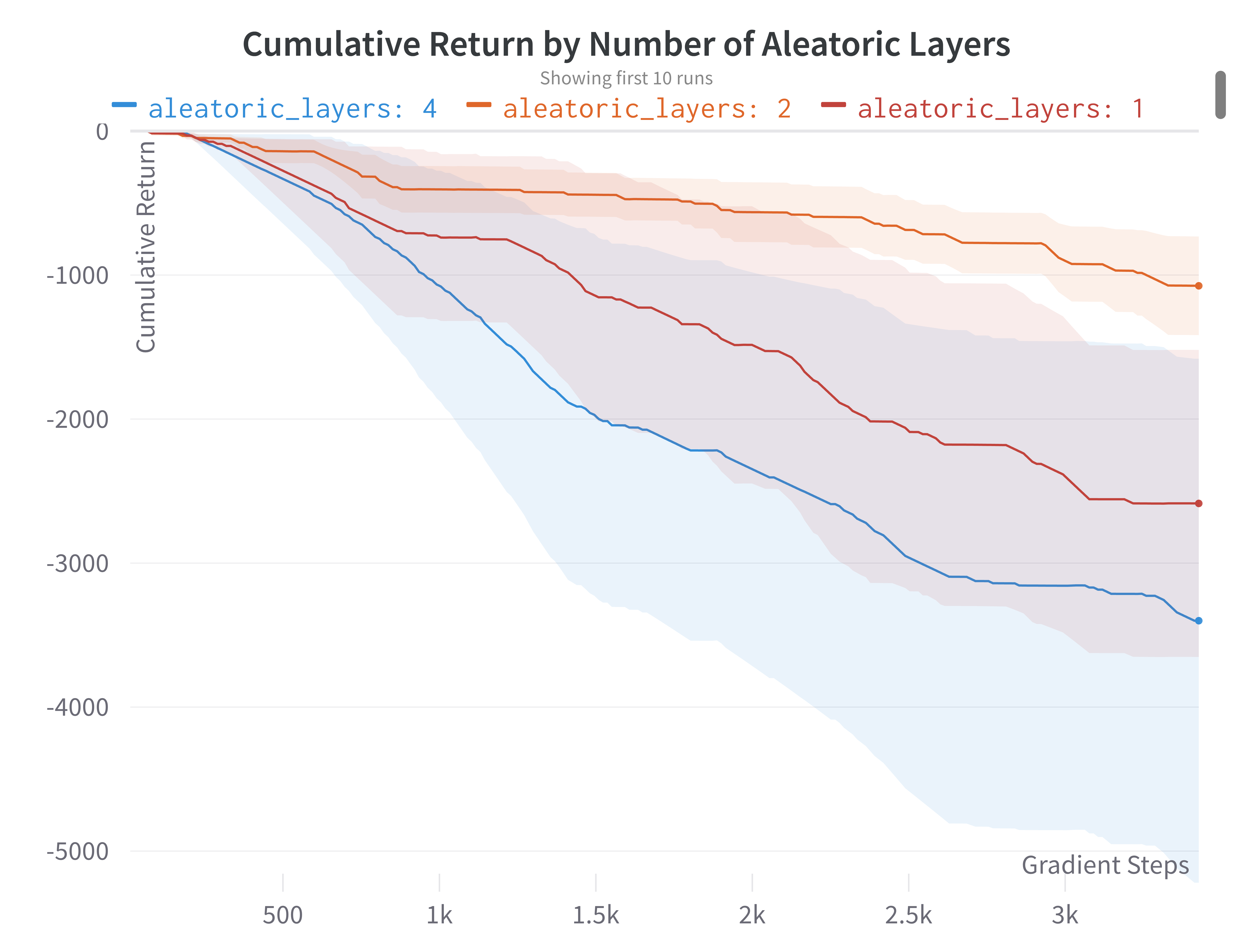}%
		\label{subfig:returns_aleatoric}%
	}\hfill
	\subfloat[Victims Saved Aleatoric Ablation]{%
		\includegraphics[width=.47\linewidth]{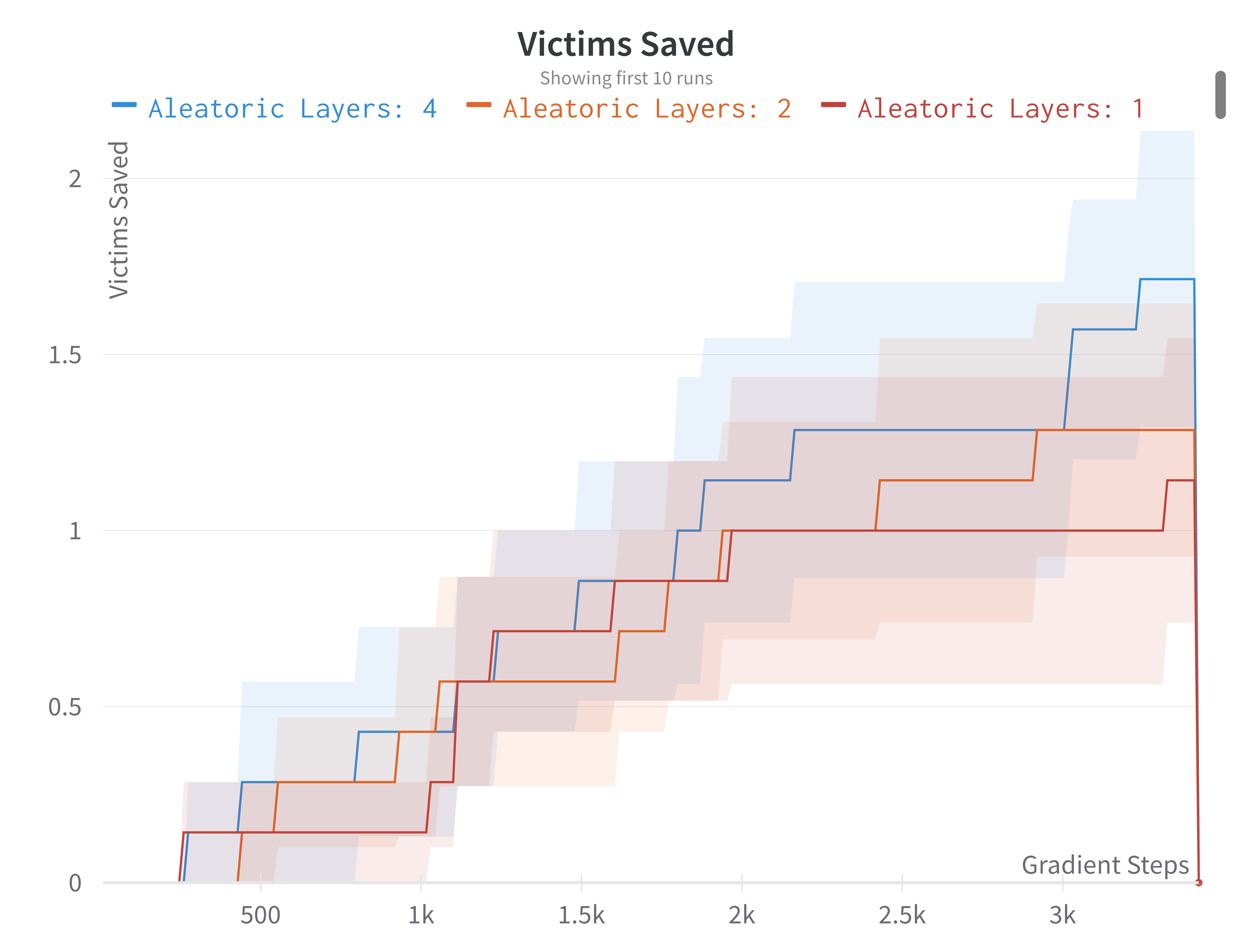}%
		\label{subfig:victims_aleatoric}%
	}
	\label{fig:aleatoric}
	\caption{Aleatoric Network Ablation}
\end{figure}

We repeat the ablation carried out in the Tiger Problem for this new domain, demonstrating the existing approaches that learn a contextual optimal policy (i.e. state of the art model-free approaches such as BBAC \citep{Fellows21} and BootDQN+Prior \citep{Osband19}) cannot succeed in this challenging setting. This corresponds to using a function approximator with no capacity to represent history. BEN provides a clear improvement over these existing methods in terms of cumulative return in \cref{fig:search_one_results}. To understand why, for each approach we plot the number of victims rescued in \cref{subfig:victims} and hazards hit in \cref{subfig:hazards}. Although both approaches save a similar number of victims, the contextual approach hits an order of 10 times more hazards than \textsc{ben}.
These results demonstrate that QBRL approaches struggle to solve this problem  whereas \textsc{ben} is slightly more conservative, yet does not hit nearly as many hazards, as we would expect given the disproportionally greater negative reward for hitting a hazard than rescuing the victim in our environment. 
\paragraph{Capacity for Representing Aleatoric Uncertainty}
We now investigate how reducing increasing the capacity of our aleatoric network affects the performance in this domain. We increase the number of aleatoric flow layers from 1 to 4 and plot the returns in \cref{subfig:returns_aleatoric} and the number of victims rescued in \cref{subfig:victims_aleatoric}.
We see that for this environment, 2 flow layers yields the best returns. As the number of aleatoric flow layers is determines the hypothesis space, our results provide evidence that there exists a trade-off between specifying a rich enough hypothesis space and a hypothesis space that is too general for the problem setting. For 4 layers, the hypothesis space is too general to learn how to behave optimally given the number of minimisation steps whereas for 1 layer, the agent cannot represent aleatoric uncertainty sufficiently to learn a policy that is useful for the environment. This ablation also supports our central claim that aleatoric uncertainty cannot be neglected in model-free Bayesian approaches.

\begin{figure}[t!]
	\subfloat[Return Prior Ablation]{%
		\includegraphics[width=.47\linewidth]{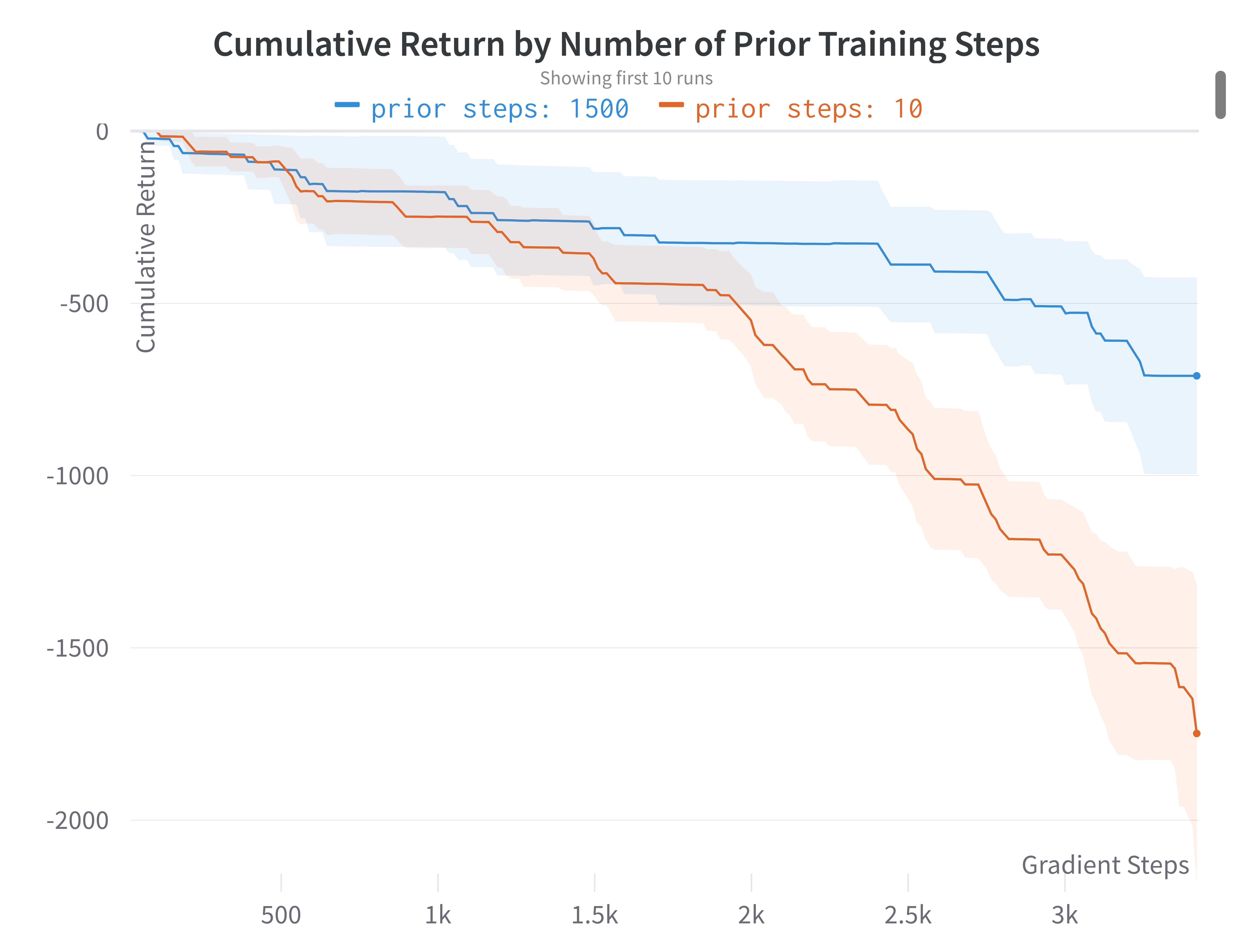}%
		\label{subfig:returns_aleatoric_prior}%
	}\hfill
	\subfloat[Victims Saved Prior Ablation]{%
		\includegraphics[width=.47\linewidth]{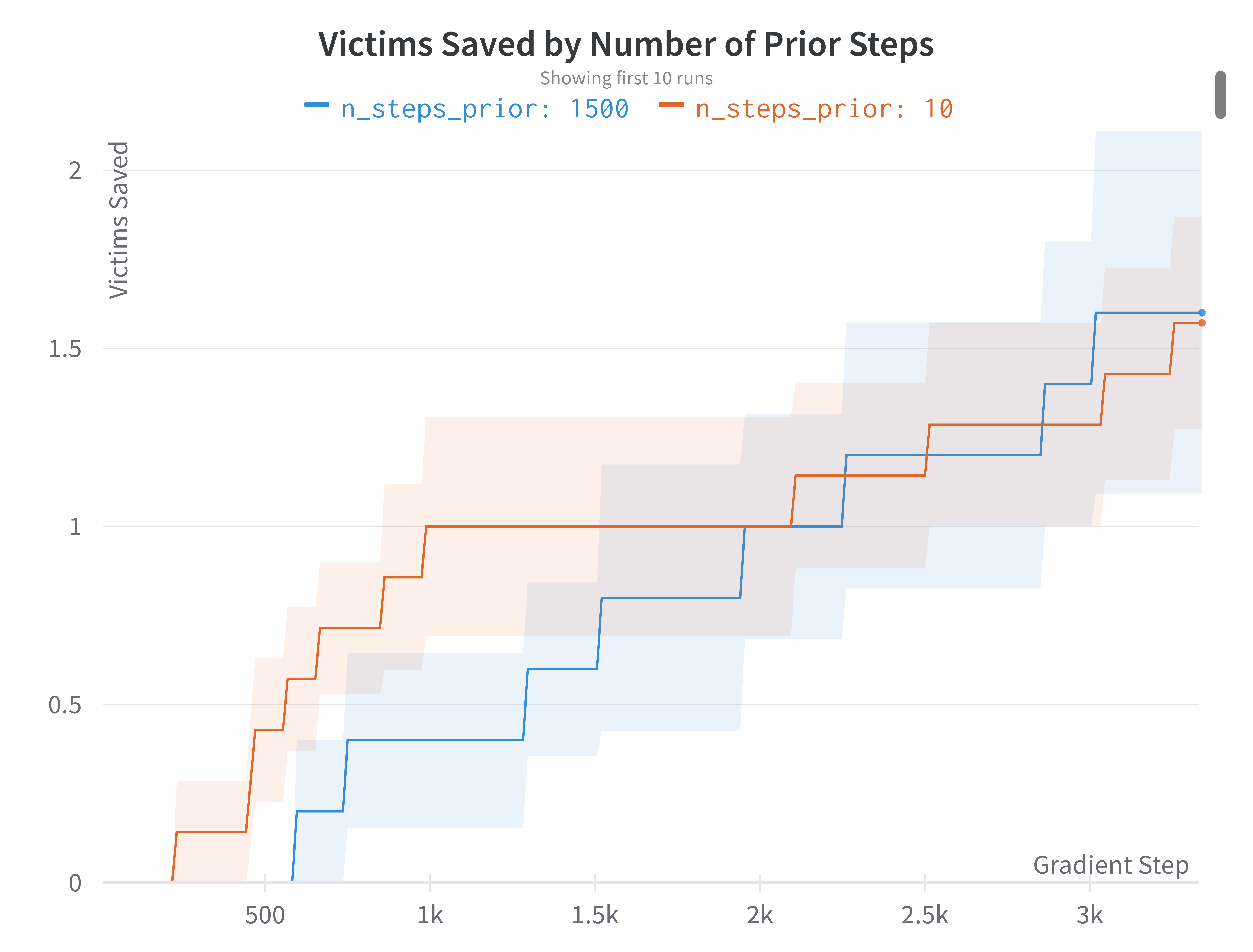}%
		\label{subfig:victims_aleatoric_prior}%
	}
	\label{fig:prior}
	\caption{Aleatoric Network Ablation}
\end{figure}

\paragraph{Incorporation of Prior Knowledge }
Finally, we investigate how our prior training regime affects the performance of \textsc{ben}, varying the number of prior gradient training steps according to \cref{alg:approxBRL_prior}. Results are plotted in 
\cref{fig:prior}. A key motivation for taking a Bayesian approach to RL is the ability to formally exploit prior knowledge. We use this ablation to demonstrate how knowledge provided by simple simulations can be incorporated into \textsc{ben}'s pre-training regime. As we decrease the number of prior pretraining MSBBE minimisation steps, we see that performance degrades in the zero-shot settling as expected. Moreover, this ablation shows that a relatively few number of pre-training steps are needed to achieve impressive performance once the agent is deployed in an unknown MDP, supporting our central claim that \textsc{ben} is computationally efficient.

\end{document}